%% file: MAS_arxiv.tex
\documentclass[english]{article}

\usepackage{geometry}
\usepackage[utf8]{inputenc}
\usepackage[T1]{fontenc}

\usepackage{amsmath}
\usepackage{amssymb}
\usepackage{amsfonts}
\usepackage{amsthm}
\usepackage{bm}
\usepackage{mathtools}
\usepackage{nicefrac}
\allowdisplaybreaks

\usepackage[sort,compress]{natbib}

\usepackage{booktabs}
\usepackage[table]{xcolor}
\usepackage{graphicx}
\usepackage{subcaption}
\usepackage{float}
\usepackage{multirow}
\usepackage{makecell}
\usepackage{caption}
\usepackage{titletoc}
\usepackage{microtype}

\usepackage{fontawesome5}
\usepackage{pgfplots}
\pgfplotsset{compat=1.18}
\usepgfplotslibrary{polar}
\usepackage{listings}
\usepackage[most]{tcolorbox}
\tcbuselibrary{listings,breakable,skins}
\usepackage{courier}
\usepackage[normalem]{ulem}
\usepackage{enumitem}

\usepackage[colorlinks=true, linkcolor=blue, citecolor=blue, urlcolor=blue]{hyperref}
\usepackage{url}
\theoremstyle{remark}
\newtheorem{remark}{Remark}

\definecolor{takeawayfill}{HTML}{F5F8FC}
\definecolor{takeawayborder}{HTML}{1F4E8C}

\definecolor{darkgreen}{RGB}{0,100,0}
\definecolor{strongpink}{RGB}{200,0,100}
\definecolor{stronggreen}{RGB}{0,140,90}

\definecolor{blueboxfill}{HTML}{F5F8FC}
\definecolor{blueboxborder}{HTML}{1F4E8C}
\definecolor{blueboxtitle}{HTML}{E6F0FA}

\tcbset{
    bluebox/.style={
        colback=blueboxfill,
        colframe=blueboxborder,
        coltitle=blueboxborder,
        colbacktitle=blueboxtitle,
        fonttitle=\bfseries,
        boxrule=1pt,
        arc=3pt,
        outer arc=3pt,
        width=\linewidth,
        boxsep=3pt,
        left=4pt,
        right=4pt,
        top=3pt,
        bottom=3pt,
        before skip=5pt,
        after skip=5pt,
        breakable,
        enhanced,
    }
}

\definecolor{grayboxfill}{HTML}{F7F7F7}
\definecolor{grayboxborder}{HTML}{8A8A8A}
\definecolor{grayboxtitle}{HTML}{EAEAEA}

\tcbset{
    graybox/.style={
        colback=grayboxfill,
        colframe=grayboxborder,
        coltitle=black,
        colbacktitle=grayboxtitle,
        fonttitle=\bfseries,
        boxrule=1pt,
        arc=3pt,
        outer arc=3pt,
        width=\linewidth,
        boxsep=3pt,
        left=4pt,
        right=4pt,
        top=3pt,
        bottom=3pt,
        before skip=5pt,
        after skip=5pt,
        breakable,
        enhanced,
    }
}

\makeatletter
\def\@fnsymbol#1{\arabic{footnote}}
\makeatother

\title{\textit{MAS-PromptBench}: When Does Prompt Optimization Improve Multi-Agent LLM Systems?}

\author{
  Juyang Bai\thanks{Department of Electrical and Computer Engineering, Johns Hopkins University, Baltimore, MD 21218, USA.} \\
  Johns Hopkins University \\
  jbai23@jh.edu\\ 
  \and
  Laixi Shi\footnotemark[1] \\
  Johns Hopkins University \\
  laixis@jhu.edu
}

\date{\today}

\begin{document}

\maketitle

\input{sections/0-abstract}

\input{sections/1-introduction}

\input{sections/2-related-works}

\input{sections/3-method}

\input{sections/4-benchmark}

\input{sections/5-result}

\input{sections/6-conclusion}

\section*{Acknowledgments}
L. Shi and J. Bai are supported in part by Mitsubishi Electric Research Laboratories (MERL).

\bibliographystyle{apalike}
\bibliography{references}

\newpage
\appendix
\input{sections/7-appendix}

\end{document}

%% file: sections/0-abstract.tex
\begin{abstract}

    Multi-agent systems (MAS) offer a scalable path forward for agentic AI, comprising multiple LLM-based agents, each assigned a system prompt and a position within a workflow that governs inter-agent coordination and output aggregation.
System prompts thus form a critical and accessible optimization surface: they specify agents' roles and behaviors, enabling system-level improvements without model finetuning.
Although prompt optimization has shown substantial potential for single LLMs, extending it to MAS poses distinct challenges, notably an exponentially growing search space. It remains unclear \emph{whether, when, and by how much prompt optimization improves MAS performance, and how sensitive such gains are to system configuration}.
In this work, we systematically study system-prompt optimization across a broad range of MAS setups varying in task, workflow, communication protocol, and team size, benchmarking two prompt optimizers that naturally extend state-of-the-art single-agent methods.
The results reveal its potential to unlock significant gains while exposing open challenges, characterizing when and how much prompt optimization helps across diverse MAS settings.

\begin{center}
\begin{tabular}{@{}ll@{}}
    $\vcenter{\hbox{\textcolor[HTML]{0077B6}{\faHome}}}$~\textcolor{black}{Website:}  \href{https://juyangbai.github.io/MAS-PromptBench/}{\textcolor{black}{https://juyangbai.github.io/MAS-PromptBench/}} \\
    $\vcenter{\hbox{\hspace{0.07em}\textcolor[HTML]{6E40C9}{\faGithub}}}$~\hspace{0.12em}\textcolor{black}{Code:}  \href{https://github.com/juyangbai/MAS-PromptBench}{\textcolor{black}{https://github.com/juyangbai/MAS-PromptBench}} \\
\end{tabular}
\end{center}

\end{abstract}

%% file: sections/1-introduction.tex
\section{Introduction}

Agentic AI, as foundation-model-based systems that autonomously plan, use tools, and interact with the real world, is rapidly transforming daily life, industry, and scientific discovery~\cite{lu2024ai,plaat2025agentic,anthropic_claude_code_2026}. As tasks evolve from human-scale problems to organization-scale challenges that are increasingly complex, open-ended, and time-sensitive, single-agent architectures face fundamental bottlenecks in expertise breadth, context length, and sequential execution~\cite{chen2024survey}. In contrast, multi-agent systems (MAS) have emerged as a highly promising paradigm for next-generation agentic AI and general superintelligence (ASI)~\cite{anthropic_claude_code_2026,genewein2026agi}, offering scalability, timeliness, and reliability through specialization and multimodality, task decomposition and parallelism, and independent cross-checks that strengthen reasoning and factual accuracy~\cite{du2023improving}.
Concretely, a MAS typically comprises multiple LLM-based agents coordinated by a harness that manages communication, task delegation, and output aggregation, with each agent assigned an {\em instruction set} and a position in a coordination workflow~\cite{anthropic_claude_code_2026}. Throughout this paper, we refer to an LLM's instruction set as its {\em system prompt}, which may be a assembled collection comprising not only the system prompt itself but also other levels of instructions. \looseness = -1

Within this MAS design space, {\em system prompts} provide a critical and accessible optimization surface: they specify each agent's role and behavior~\cite{anthropic_system_prompts,openai_model_spec_2025,google_gemini_text_generation,meta_llama4_model_card}, enabling system-level improvement without model fine-tuning.
System prompts are among the most accessible levers available to practitioners, who often inherit a fixed configuration and seek improvements without redesigning the underlying architecture; many real-world deployments further preclude topology changes due to safety, compliance, or auditability constraints~\cite{hong2026maspob}.
Given the important role of system prompts, automatic prompt optimization has been studied extensively in the single-agent regime, with strong demonstrated benefits~\cite{zhou2022large,yang2024large,wang2024promptagent,khattab2023dspy,agrawal2025gepa}.

Whether such  gains transfer to the multi-agent setting remains underexplored. Extending prompt optimization to MAS introduces qualitatively new challenges: inter-agent prompt dependencies, compounded by coordination dynamics across multi-turn interactions, induce a combinatorial search space that grows exponentially with the number of agents. As illustrated in Figure~\ref{fig:intro}, the current effect of prompt optimization on MAS varies dramatically across tasks and topologies---ranging from substantial gains to equally severe performance drops. Meanwhile, most influential MAS frameworks---including AutoGen, CrewAI, CAMEL, MetaGPT, ChatDev, and AgentVerse~\cite{wu2024autogen,crewai_repo,li2023camel,hong2024metagpt,qian2024chatdev,chen2024agentverse}, as well as collaboration workflows such as debate~\cite{du2023improving,liang2024encouraging}---still rely on manually crafted system prompts. Recent works have begun to address this gap, either by developing dedicated MAS prompt optimization algorithms~\cite{xia2026hivemind,shen2025optimizing,zhang2026mapro,hong2026maspob} or by jointly optimizing prompts alongside orchestration components such as workflow topology~\cite{zhao2025connecting,zhou2025multi,hu2025automated}.
Yet these works evaluate on different tasks (e.g., math, coding, stock trading), configurations, and baselines, making cross-comparison difficult and leaving a fundamental question open:
\begin{center}
\emph{How much can prompt optimization help in MAS, and how does its effect vary across configurations?}
\end{center}

We address this through a benchmark-driven study that quantifies the gains from system-prompt optimization across a diverse set of fixed MAS configurations, pinpointing the regimes where prompt optimization offers substantial untapped improvement room and regimes where MAS sensitivity calls for more principled algorithm design.
These findings provide a roadmap for designing future prompt optimization algorithms and, more broadly, for MAS design, where prompts are tightly coupled with other configuration choices.
Our main contributions are as follows:

\begin{figure}[t]
    \centering
    \includegraphics[width=\linewidth]{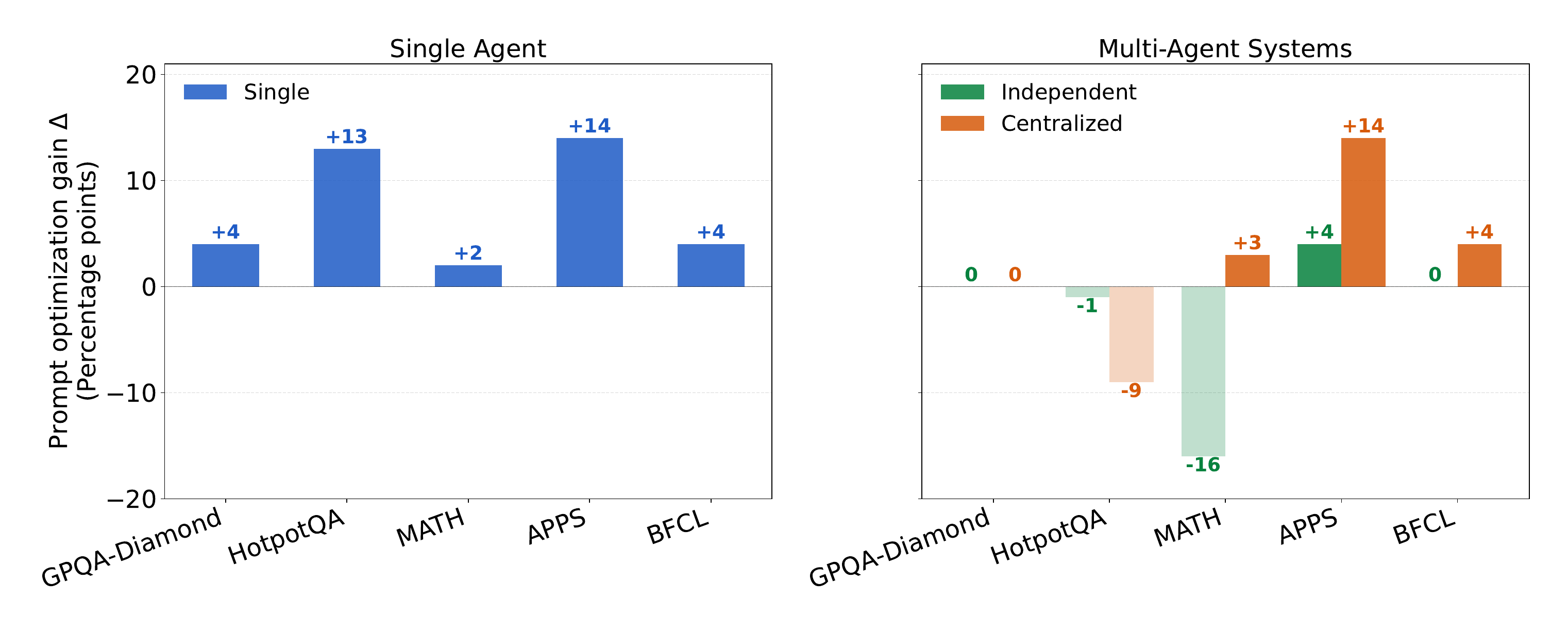}
    \caption{Prompt-optimization gains using a state-of-the-art optimizer GEPA in single-agent and multi-agent settings. While GEPA consistently improves single-agent performance across all five diverse tasks, its natural multi-agent extension yields highly variable effects across tasks and workflow topologies, ranging from large gains to severe performance drops.
    }
    \label{fig:intro}
\end{figure}

\begin{itemize}
    \item \textbf{MAS-PromptBench: a benchmark for MAS prompt optimization.} We introduce a comprehensive benchmark for evaluating system-prompt optimizers for MAS. It spans diverse MAS configurations across task domains (reasoning, coding, and tool calling), five workflow topologies (comprising both existing and newly constructed systems), communication protocols ranging from free-form to highly structured coordination, varying team sizes, and two default prompt optimizers. The benchmark provides a foundation for proposing, analyzing, and comparing system-prompt optimization algorithms under controlled MAS configurations.
    \item \textbf{Prompt optimization gains and failures for MAS.} Using this benchmark, we systematically evaluate the performance gains achieved by a natural multi-agent extension of GEPA~\cite{agrawal2025gepa}, a state-of-the-art single-agent prompt optimizer, relative to default system prompts. The results highlight the promise of prompt optimization for MAS: improvements reach up to 24.0 percentage points. Yet they also reveal the need for principled algorithms tailored to multi-agent settings, as performance can drop by as much as 16.0 percentage points for certain configurations.
    \item \textbf{Insights into when prompt optimization works for MAS.} Prompt optimization shows greater potential when tasks have explicit, controllable, and verifiable agent-local behaviors, and when communication protocols impose an explicit shared structure that makes agent interactions easier to control and transfer; it also needs to be workflow-topology-aware. In addition, prompt optimization becomes harder as team size grows, confirming the challenges of scaling MAS prompt optimization and motivating principled and more scalable, robust algorithms.
\end{itemize}

%% file: sections/2-related-works.tex
\section{Related Work}
\label{sec:related}

\paragraph{Prompt Optimization for Single LLM.}

Prompt optimization improves LLM performance without updating model weights; see \cite{ramnath2025systematic,chang2024efficient} for detailed reviews. Prompts are typically categorized into system (hard) prompts as discrete text instructions and soft prompts as continuous embeddings~\cite{chang2024efficient}. While both have shown effectiveness in single-agent settings, we focus on system prompts---the discrete instructions that specify each agent's role---due to their interpretability and direct role in specifying agent behavior in MAS.

Most system prompt optimization methods can be viewed as searching over a discrete instruction space~\cite{chang2024efficient}. Existing approaches fall into three categories: (1) sampling-based methods that generate and select candidate prompts using task feedback, including self-generated methods~\cite{wang2023self}, LLM-as-optimizer approaches such as APE~\cite{zhou2022large} and OPRO~\cite{yang2024large}, planning-based methods such as PromptAgent~\cite{wang2024promptagent}, and evolutionary methods such as EvoPrompt~\cite{guo2024connecting} and PromptBreeder~\cite{fernando2023promptbreeder}; (2) feedback-based methods that leverage directional signals such as reinforcement-learning rewards~\cite{deng2022rlprompt}, textual gradients~\cite{pryzant2023automatic,yuksekgonul2024textgrad}, or self-reflection~\cite{madaan2023self,shinn2023reflexion}; and (3) editing-based methods that refine prompts through local operations such as insertion, deletion, or paraphrasing~\cite{prasad2023grips}. These techniques are also integrated into broader frameworks such as DSPy~\cite{khattab2023dspy}, which optimizes instructions within multi-stage LLM programs via algorithms such as MIPROv2~\cite{opsahl2024optimizing}. In this work, instead of single-agent, we focus on prompt optimization for multi-agent LLM systems, where it remains unclear whether single-agent gains transfer. To investigate, we evaluate when and how much prompt optimization improves MAS performance across a broad range of setups varying in task, workflow, communication protocol, team size, and different prompt optimizers.

\paragraph{Prompt Optimization in Multi-Agent LLM Systems.}

Many influential MAS still rely on manually designed system prompts, specifying roles in CAMEL, MetaGPT and ChatDev \cite{li2023camel,hong2024metagpt,qian2024chatdev}; specifying conversations and coordination patterns in AutoGen, CrewAI, and AgentVerse\cite{wu2024autogen, crewai_repo, chen2024agentverse}; instantiating collaboration patterns, such as debate~\cite{du2023improving, liang2024encouraging}, mixture-style aggregation~\cite{wang2025mixture}, or consensus among diverse LLMs~\cite{chen2024reconcile}. Recently, automatic prompt optimization for MAS has attracted growing attention and shown substantial progress. Many existing methods generate new prompts from diverse feedback sources, including failure attributes and identified underperforming agents~\cite{shen2025optimizing,li2026unifying}, as well as multi-resolution signals spanning the agent itself, its neighborhood, and global contexts~\cite{wang2026maspo}; while other work searches over a fixed, pre-generated set of candidate prompts rather than generating new ones~\cite{hong2026maspob}. Beyond these directions, researchers have also studied MAS prompt optimization for domain-specific problems~\cite{xia2026hivemind} and incorporated richer MAS-configuration information into the optimization process~\cite{zhang2026mapro}. Recognizing that system prompts are tightly coupled with other design choices, another line of work jointly optimizes prompts with orchestration components such as communication topology and hierarchical planning~\cite{zhao2025connecting,zhou2025multi}.

In this work, we ask a more foundational question: \emph{within a given MAS configuration, how much headroom does system prompt optimization actually offer, and how does this headroom vary across configurations?} We address this through a benchmark-driven study that quantifies how much system-prompt optimization helps across a diverse set of fixed MAS configurations. Specifically, we evaluate the gains—relative to default prompts—of a natural multi-agent extension of GEPA~\cite{agrawal2025gepa}, a state-of-the-art single-agent prompt optimization method. We view these findings as an empirical roadmap that can guide the design of future prompt optimization algorithms—and, more broadly, the design of MAS itself.

\paragraph{Benchmarks for multi-agent LLM systems or prompt optimization.}
Existing work on evaluating and analyzing LLM-based multi-agent systems (MAS) spans three complementary perspectives: task benchmarks, diagnostic tools, and studies of MAS design choices. First, general task benchmarks provide broad testbeds for agents' different abilities within a MAS. Examples include MultiAgentBench~\cite{zhu2025multiagentbench} for coordination quality, BFCL~\cite{patil2025berkeley} for function calling, AppWorld~\cite{trivedi2024appworld} for interactive app operation, GAIA~\cite{mialon2024gaia} for general assistant tasks, TravelPlanner~\cite{xie2024travelplanner} for multi-constraint planning, and SWE-bench~\cite{jimenez2024swe} for software repair. Second, diagnostic and debugging work addresses the interpretability gap by examining why MAS executions succeed or fail. MAST proposes a taxonomy of failure modes with an annotated trace dataset~\cite{cemri2026multi}; AutoGen Studio and AGDebugger enable interactive inspection and steering of multi-agent conversations~\cite{dibia2024autogen,epperson2025interactive}; and failure-attribution work identifies which agent and step caused a task failure~\cite{zhang2025agent}. Third, a line of work studies how different MAS configuration choices influence overall performance, including workflow topology~\cite{kim2025towards,shen2025understanding}, agent diversity~\cite{yang2026understanding}, and team size~\cite{kim2025towards,qian2025scaling}. We follow this third line, but focus on a critical yet underexplored component—the system prompt—studying how much optimizing it can improve MAS performance under a fixed surrounding configuration. Furthermore, although extensive benchmarks exist for evaluating prompt optimization in single-agent setting~\cite{zhu2024promptbench}, no comparable benchmark is available for MAS yet. To fill this gap, we introduce a systematic benchmark MAS-PromptBench that measures the gains from MAS prompt optimization across a diverse set of configurations, encompassing different tasks, workflow topologies, team sizes, and communication protocols.

%% file: sections/3-method.tex
\section{Prompt Optimization for Multi-Agent LLM Systems}
\label{sec:method}

\begin{figure}[t]
    \centering
    \includegraphics[width=\linewidth]{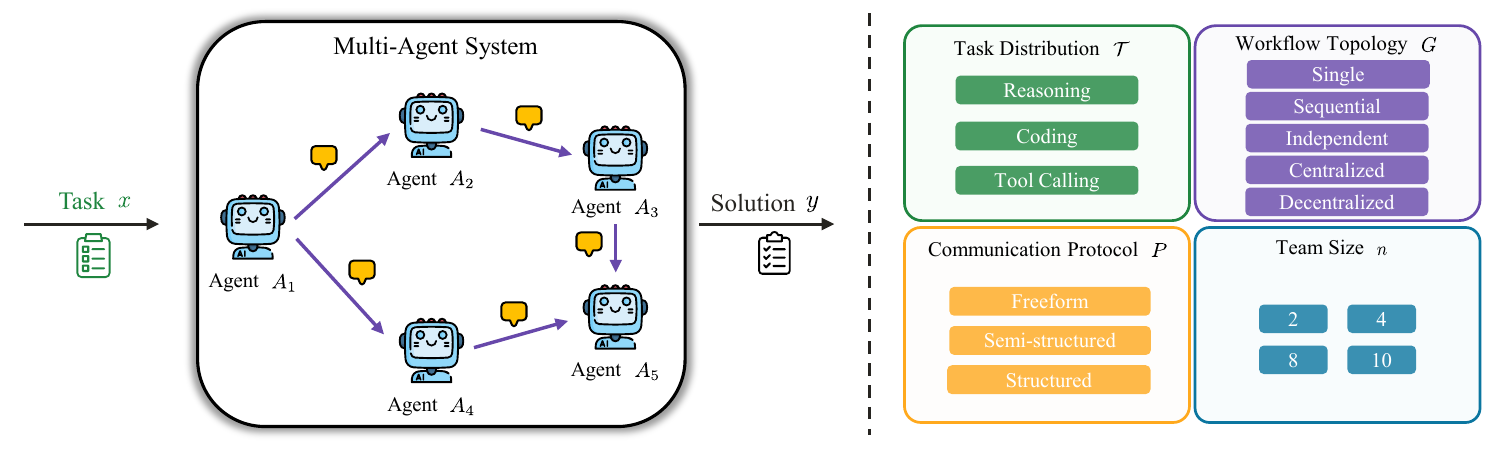}
    \caption{Overview of benchmark MAS-PromptBench. Given an input task, a multi-agent system produces a final solution through interactions among LLM-based agents. MAS-PromptBench measures prompt-optimization gains across four axes: task distribution, workflow topology, communication protocol, and team size.}
    \label{fig:protocol}
\end{figure}

\paragraph{Prompt optimization for multi-agent systems (MAS).}
We consider an LLM-based multi-agent system (MAS) \cite{khattab2023dspy, opsahl2024optimizing, agrawal2025gepa}, illustrated in Figure~\ref{fig:protocol}, represented as the tuple
\begin{equation}
\mathcal{M} = (\mathcal{A},\, G,\, P),
\end{equation}
where $\mathcal{A} = \{A_1, \dots, A_n\}$ is an ordered collection of $n$ agents, $G$ denotes the inter-agent coordination workflow, and $P$ denotes the communication protocol between agents. Each agent $A_i = (\theta_i, \pi_i)$ consists of an LLM with model parameters $\theta_i$ and a learnable system prompt $\pi_i$. 
We denote $\pi = \{\pi_1, \dots, \pi_n\}$ as the joint system prompts for all agents. 
Each task to be solved is drawn as $(x, e) \sim \mathcal{T}$ from a distribution $\mathcal{T}$, where $x \in \mathcal{X}$ is the task input and $e$ is a reference used for evaluation, such as a ground-truth answer or code unit tests. We let $\mathcal{M}(\,\cdot\,;\pi)$ denote the output function of the MAS $\mathcal{M}$ induced by the joint prompt $\pi$, with the model parameters $\theta$ held fixed. System-prompt optimization for a MAS is then formulated as the following optimization problem:
\begin{equation}
    \max_{\pi}\; \mathbb{E}_{(x,e) \sim \mathcal{T}}\!\left[\mu\!\left(\mathcal{M}(x;\pi), e\right)\right] \quad \text{s.t.}\;\; \ell_{\mathsf{rollouts}} \le B,
    \label{eq:opt}
\end{equation}
where the metric $\mu:\mathcal{Y}\times\mathcal{E}\to[0,1]$ evaluates the output
$y=\mathcal{M}(x;\pi)$ against the reference $e$,
and $\ell_{\mathsf{rollouts}}$ denotes the number of MAS execution rounds, each comprising one MAS run to solve a task followed by an evaluation of its output.

\begin{remark}[{\em System prompt} as an umbrella term for the instruction set]
Within a multi-agent system (MAS), each agent is typically prompted with a carefully assembled input package determined by the harness, which may include the user's request, a layered set of instructions, task-specific skills, relevant memory, retrieved context, and available tools. We focus on the instruction set, which typically comprises four components that differ in authority, scope, and purpose, forming a hierarchy that guides behavior from global constraints to task-specific procedures. At the top, system prompts define universal rules, such as safety and tool-use protocols; developer or product instructions then specify application-level operation, including the agent's role and workflow; project instructions specify task-level conventions, such as coding standards and testing procedures; and finally, skills provide reusable guidance for particular task classes, such as data analysis or code modification. Throughout this paper, we use {\em system prompt} as an umbrella term for the instruction set excluding skills, without distinguishing among them. In practice, agentic systems may employ multiple instruction layers; our formulation extends naturally to optimizing an individual component or combination thereof, which we leave to future work.

\end{remark}

\paragraph{Metric: prompt optimization gains in MAS.} In this work, we will conduct systematic study of whether, when, and to what extent system-prompt optimization improves MAS performance across different MAS configurations. To this end, we investigate diverse MAS configurations determined by four critical components $(\mathcal{T}, G, n, P)$: the task distribution $\mathcal{T}$, the coordination workflow or topology $G$, the team size $n$, and the communication protocol $P$. Any prompt optimizer can, in principle, be used to optimize prompts by solving \eqref{eq:opt}. In this work, we primarily focus on two natural multi-agent extension of state-of-the-art single-agent prompt optimization methods: GEPA~\cite{agrawal2025gepa} and MIPRO~\cite{opsahl2024optimizing}
Let $\pi^{0}$ denote the initialized (unoptimized) system prompt and $\pi^\star$ the optimized prompt obtained by solving \eqref{eq:opt}. Fixing an optimizer for solving \eqref{eq:opt} and the base model, the \textit{prompt-optimization gain} for the MAS with a configuration $(\mathcal{T}, G, n, P)$ is then defined as
\begin{equation}
    \Delta(\mathcal{T}, G, n, P) \;:=\; \mathbb{E}_{(x,m) \sim \mathcal{T}}\!\bigl[\mu\!\left(\mathcal{M}(x;\pi^\star), y\right) \;-\; \mu\!\left(\mathcal{M}(x;\pi^0), y\right)\bigr].
    \label{eq:delta}
\end{equation}

%% file: sections/4-benchmark.tex
\section{MAS-PromptBench: Prompt Optimization for MAS Benchmark}
\label{bench}

While benchmarks and evaluation protocols exist for single-agent prompt optimization, comparable resources for multi-agent systems remain underdeveloped. We fill this gap by introducing a benchmark designed to support extensive and controlled investigation for prompt optimization across diverse MAS configurations, summarized in Table~\ref{tab:config} and Figure~\ref{fig:protocol}.

\input{tables/table-config-matrix.tex}

The proposed benchmark evaluates the prompt-optimization gain using the metric in \eqref{eq:delta}, which compares the optimized prompts $\pi^\star$ against the initialized prompts $\pi_0$ for a fixed prompt optimizer and MAS configuration $(\mathcal{T}, G, n, P)$. This controlled protocol makes the benchmark useful in at least two ways. First, given any system-prompt optimizer, it provides an extensive testbed across diverse MAS configurations, enabling direct evaluation of how optimization gains vary with task, topology, communication structure, and team size. Second, its modular design supports controlled, component-level studies: one can vary a single MAS factor—such as the optimizer, topology, protocol, or team size—while holding all others fixed to isolate its effect on system-level performance. The benchmark is also flexible and extensible across all of these components, as new tasks or MAS configurations can be introduced as additional configuration values, allowing it to be tailored to different application domains and user requirements. We primarily provide two optimizers---multi-agent extensions of state-of-the-art single-agent prompt optimization methods, GEPA~\cite{agrawal2025gepa} and MIPRO~\cite{opsahl2024optimizing}---which we call MAS-GEPA and MAS-MIPRO. This benchmark evaluates each task dataset using its official evaluation protocol; detailed metric definitions and implementations are provided in Appendix~\ref{app:tasks}.

%% file: tables/table-config-matrix.tex
\begin{table}[H]
    \caption{Overview of the modular configuration of MAS-PromptBench.}
    
    \centering
    \small
    \renewcommand{\arraystretch}{1.25}
    \begin{tabular}{llp{0.62\linewidth}}
        \toprule
        \textbf{Factor} & \textbf{\#} & \textbf{Details} \\
        \midrule
        Framework    & 4 & LangGraph~\cite{langgraph_repo}, CrewAI~\cite{crewai_repo}, AutoGen~\cite{wu2024autogen}, OpenAI Agents SDK~\cite{openai_agents_evolution_2026} \\
        Task          & 9 & \textit{Reasoning} (3): GPQA-Diamond~\cite{rein2023gpqa}, HotpotQA~\cite{yang2018hotpotqa}, MATH~\cite{hendrycks2021math}; \textit{Coding} (3): LiveCodeBench~\cite{jain2025livecodebench}, APPS~\cite{hendrycks2021apps}, SWE-Bench Verified~\cite{jimenez2024swe}; \textit{Tool-calling} (3): BFCL~\cite{patil2025berkeley}, ToolHop~\cite{ye2025toolhop}, API-Bank~\cite{li2023api} \\
        Topology      & 5 & Single, Independent, Sequential, Centralized, Decentralized \\
        Communication & 3 & Freeform, Semi-structured, Structured \\
        Team size     & 4 & $n \in \{2, 4, 8, 10\}$ \\
        Optimizer    & 2 & MAS-GEPA~\cite{agrawal2025gepa}, MAS-MIPRO~\cite{opsahl2024optimizing} \\
        \bottomrule
    \end{tabular}
    \label{tab:config}
\end{table}

%% file: sections/5-result.tex
\section{Empirical Study of Prompt Optimization in MAS}
\label{res}

Armed with the MAS-PromptBench 
benchmark, in this section we conduct a systematic study to answer: \emph{How much can prompt optimization help in MAS, and how does its effect vary across configurations?} 
We subsequently investigate the four critical MAS configuration factors: task (Sec.~\ref{subsec:task}), workflow topology (Sec.~\ref{subsec:topology}), communication protocol (Sec.~\ref{subsec:comm}), and team size (Sec.~\ref{subsec:team-size}). We mainly use the natural multi-agent extension of GEPA (named MAS-GEPA) as the prompt optimizer for Sec.~\ref{subsec:task}-Sec.~\ref{subsec:team-size}, with an ablation study of another prompt optimizer adapted from MIPRO (named MAS-MIPRO) in Sec.~\ref{sec:ablation}. Both MAS-GEPA and MAS-MIPRO optimize each agent’s system prompt separately and sequentially, using feedback from the overall MAS execution evaluation and the agent’s own experience traces. Details are provided in Appendix~\ref{app:optimziers}.

\subsection{Task}
\label{subsec:task}

We first study how prompt-optimization gains vary across task domains that spans nine tasks across three domains: reasoning, coding, and tool-calling. In this subsection on tasks, we evaluate a range of popular existing MAS frameworks with naturally differing topologies, shown in Table~\ref{tab:tab-task}; in all remaining studies on MAS configurations, we instead use the LangGraph framework to construct the different configurations, for flexibility and fairness.
Table~\ref{tab:tab-task} shows that system-prompt optimization is broadly promising across diverse tasks: averaged over topologies, it improves performance on seven out of nine tasks, with the largest average gain of $+10.0\%$ points on APPS. Individual MAS configurations show even larger gains: Sequential in BFCL improves by $+24.0$ points, and Sequential in APPS improves by $+18.0$ points.

The gains are larger and more consistent for coding and tool-calling tasks than for reasoning tasks. At the task level, the maximum average gains for coding and tool-calling are $+10.0$ points on APPS and $+6.4$ points on BFCL, respectively, whereas the largest average gain among reasoning tasks is only $+3.2$ points on HotpotQA. At the topology-configuration level, coding and tool-calling tasks achieve maximum gains of $+18.0$ points on Sequential APPS and $+24.0$ points on Sequential BFCL, whereas reasoning tasks reach only $+8.0$ points on Sequential MATH. The same trend holds at the domain level in average: coding benchmarks improve by $+3.7$ points on average and tool-calling benchmarks by $+4.3$ points, compared with only $+1.3$ points for reasoning benchmarks.

We hypothesize that the difference arises from the extent to which each task can be decomposed into an explicit routine with controllable local behaviors. Coding tasks expose verifiable artifacts—small program steps can be checked by compilation and tests, while tool-calling tasks offer structured and clear interfaces through explicit function names and outcome formats that system prompts can directly shape. Such behaviors propagate through the MAS workflow with little ambiguity, allowing local prompt improvements to survive downstream coordination. Reasoning tasks, in contrast, rely on correlated logical steps with implicit intermediate feedback, so local improvements or errors are often discarded, overwritten, or amplified before reaching the final answer. Prompt optimization is thus most effective when tasks provide structured interfaces through which agent-level changes can be clearly controlled, preserved, and transferred across agents.

\begin{tcolorbox}[
    colback=takeawayfill,
    colframe=takeawayborder,
    boxrule=1.2pt,
    arc=3pt,
    left=6pt, right=6pt, top=6pt, bottom=6pt]

    \textbf{\textit{\faLightbulb\ Takeaway.}} 
Prompt optimization shows greater potential on tasks with explicit, controllable, and verifiable agent-local behaviors, such as coding and tool-calling, than on reasoning tasks.

\end{tcolorbox}

\input{tables/table-task-cal}

\subsection{Workflow Topology}
\label{subsec:topology}

\begin{figure}[t]
    \centering
    \includegraphics[width=\linewidth]{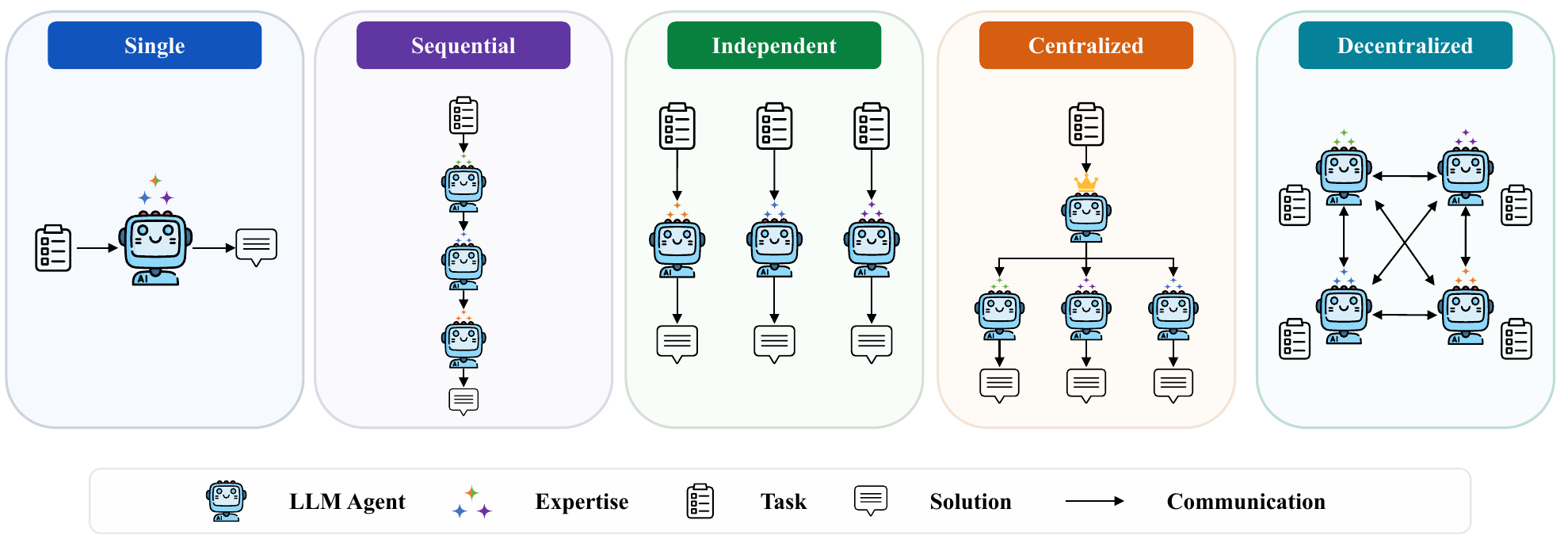}
    \caption{The five coordination structures evaluated by our protocol. \textit{Single} is the single-agent baseline. \textit{Independent} uses $n$ parallel agents whose outputs are aggregated without inter-agent messaging. \textit{Sequential} forms a directed chain $A_1 \to A_2 \to \cdots \to A_n$ with no backward edges. \textit{Centralized} uses a coordinator to route subtasks to workers that do not communicate with one another. \textit{Decentralized} allows all agents to exchange messages over a fully connected graph for a fixed number of rounds. Arrows indicate message flow; nodes indicate agents.}
    \label{fig:topologies}

    \vspace{1em}

\input{tables/table-topo-cal}

\end{figure}

A workflow topology refers to the inter-agent coordination graph $G$, which determines how agent outputs (\textit{messages}) are routed, combined, and exposed to other agents en route to the final outcome. We again use the natural multi-agent extension of GEPA as the prompt optimizer. To study the room for improvement and the level of difficulty across diverse MAS topologies, we evaluate prompt-optimization gains under the following four multi-agent topologies along with a single-agent baseline, as illustrated in Figure~\ref{fig:topologies}.
\begin{itemize}
    \item {\em Single:} A single LLM serves as the baseline.
    \item {\em Independent:} $n$ agents solve the task in parallel without inter-agent communication, and their outputs are aggregated by majority vote.
    \item {\em Sequential:} Agents form a directed chain $A_1 \to A_2 \to \cdots \to A_n$ with no backward edges; each agent receives the previous agent's output as input toward the final answer.
    \item {\em Centralized:} A coordinator dispatches subtasks to sub-agents $A_1, \dots, A_n$, collects their outputs, and aggregates them into the final answer; sub-agents do not communicate with one another throughout the process. 
    \item {\em Decentralized:} All $n$ agents communicate over a fully connected graph and exchange messages once, after which their final-round outputs are aggregated by majority vote for question-answering tasks or best-of-$N$ test-pass for coding tasks.
\end{itemize}

As shown in Table~\ref{tab:tab-topo}, the average prompt-optimization gains across the four MAS topologies (with a maximum of $+2.3$ points) are all smaller than that of the single-agent baseline at $+4.2$ points, indicating that MAS poses substantially greater challenges for prompt optimization. Moreover, for a fixed optimizer, gains vary considerably across topologies: on API-Bank, optimization improves the Sequential topology by $+9.0$ points but degrades the Centralized topology by $-5.0$; on SWE-Bench Verified, it degrades the Independent topology by $-3.3$ points yet improves the Centralized topology by $+3.3$. This disparity motivates topology-aware, tailored prompt optimization approaches rather than a one-size-fits-all method. In particular, optimization on the Independent topology can even hurt performance—dropping by $-16.0$ points on MATH and by $-0.5$ points on average—suggesting that uncoordinated prompt revisions across parallel agents may erase one another's gains. The Centralized topology, in contrast, tends to amplify both successes and failures relative to other topologies, improving APPS by $+14.0$ points while hurting HotpotQA by $-9.0$.

\begin{tcolorbox}[
    colback=takeawayfill,
    colframe=takeawayborder,
    boxrule=1.2pt,
    arc=3pt,
    left=6pt, right=6pt, top=6pt, bottom=6pt]

    \textbf{\textit{\faLightbulb\ Takeaway.}} Multi-agent systems need topology-aware prompt optimizers.
    
\end{tcolorbox}

\subsection{Communication Protocol}
\label{subsec:comm}

\begin{figure}[t]
    \centering
    \includegraphics[width=\linewidth]{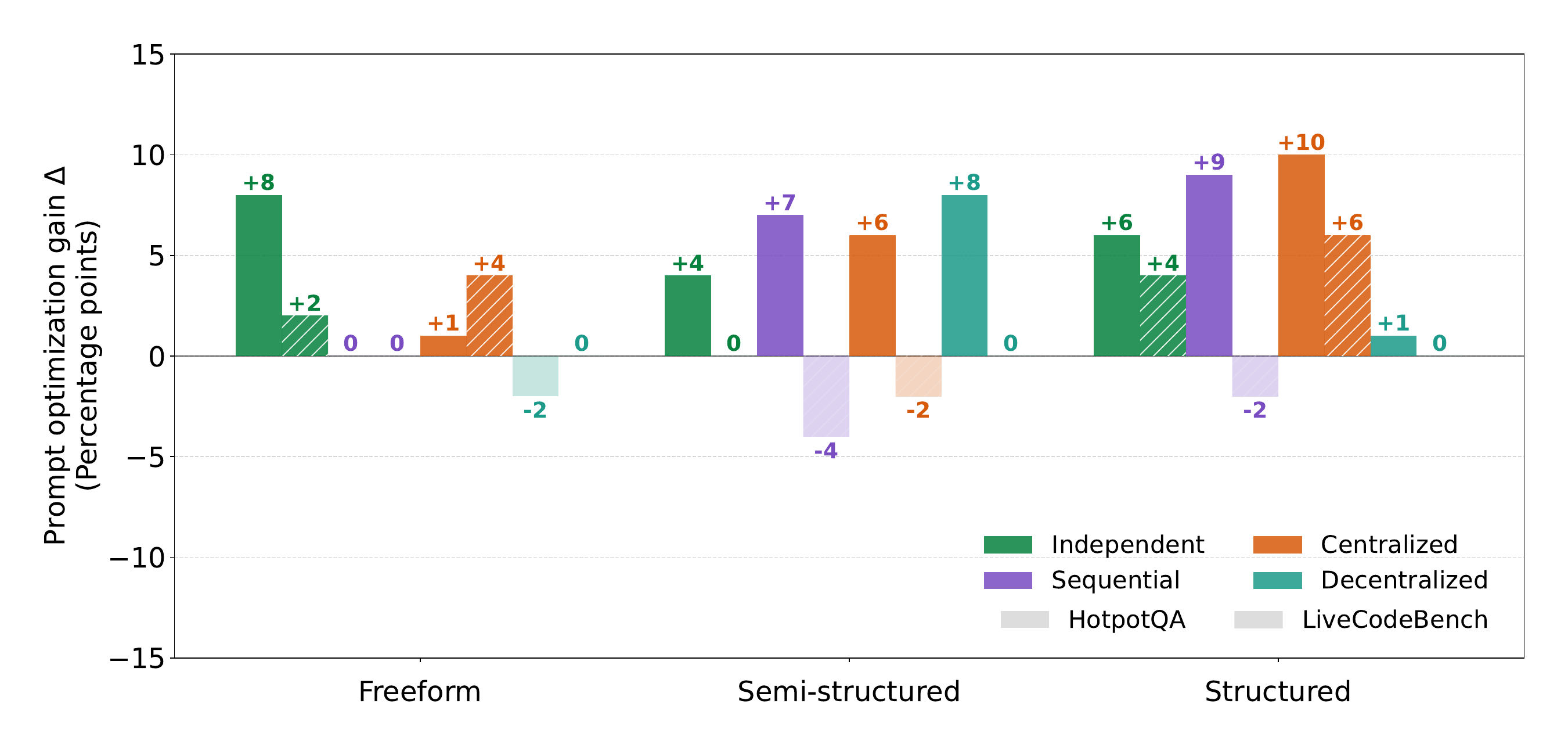}
    \caption{Prompt-optimization gains of MAS-GEPA across diverse communication protocols: Freeform, Semi-structured, and Structured, on HotpotQA and LiveCodeBench. More structured protocols give MAS prompt optimization more room to improve.
    }
    \label{fig:gepa}

\end{figure}

A communication protocol specifies the format of inter-agent messages. Since downstream agents observe only the information explicitly written by upstream agents, an underspecified or overly redundant protocol may obscure salient information or direct attention to irrelevant details. To study how communication structure affects prompt optimization, we consider three protocols with increasing levels of structure;concrete examples of each protocol are provided in Appendix~\ref{app:communication-formats}:

\begin{itemize}
    \item {\em Freeform:} Agents exchange unrestricted natural-language messages with no required fields or templates. This protocol gives agents maximum flexibility, but downstream agents must infer which information is most relevant.
    
    \item {\em Semi-structured:} Agents communicate through a small set of prescribed slots that summarize the sender's status, evidence, confidence, and intended next step. Each slot is still filled in natural language, making the message easier to scan while preserving flexibility for task-specific details.
    \item {\em Structured:} Agents communicate using a JSON-style format with a fixed set of predefined slots for critical information, such as status, summary, confidence level, supporting evidence, and next action. Unlike the semi-structured protocol, each slot's value follows a more constrained format drawn from a predefined, finite set of options. This makes message organization more consistent and reduces ambiguity across agents, but also limits how freely agents can express task-specific details.
\end{itemize}

The overall results are reported in Table~\ref{tab:tab-comm}, and Figure~\ref{fig:gepa} shows that more structured communication protocols yield consistently larger prompt-optimization gains. The average gain increases from $+1.6$ points under Freeform messages to $+2.4$ under Semi-structured messages and $+4.3$ under Structured messages.  The gains are largest on HotpotQA, a multi-hop question-answering task where answering often requires combining evidence from multiple Wikipedia passages. Here, agents must collect, preserve, and pass intermediate evidence across reasoning steps, making efficient communication essential for downstream agents to interpret and use upstream outputs.
In contrast, gains are smaller and less consistent on LiveCodeBench, where code correctness is ultimately determined by executable code and test outcomes, making performance less sensitive to message format once the code artifact is produced. Overall, prompt optimization is most effective when communication protocols provide a shared structure that makes agent state, evidence, confidence, and requests explicit, allowing local prompt improvements to propagate more reliably through the MAS workflow.

\begin{tcolorbox}[
    colback=takeawayfill,
    colframe=takeawayborder,
    boxrule=1.2pt,
    arc=3pt,
    left=6pt, right=6pt, top=6pt, bottom=6pt]

    \textbf{\textit{\faLightbulb\ Takeaway.}} Communication protocols with explicit shared structure makes agent interactions easier to control and transfer, giving MAS prompt optimization more room to improve.

\end{tcolorbox}

\subsection{Team Size}
\label{subsec:team-size}

\begin{figure}[t]
    \centering
    \includegraphics[width=\linewidth]{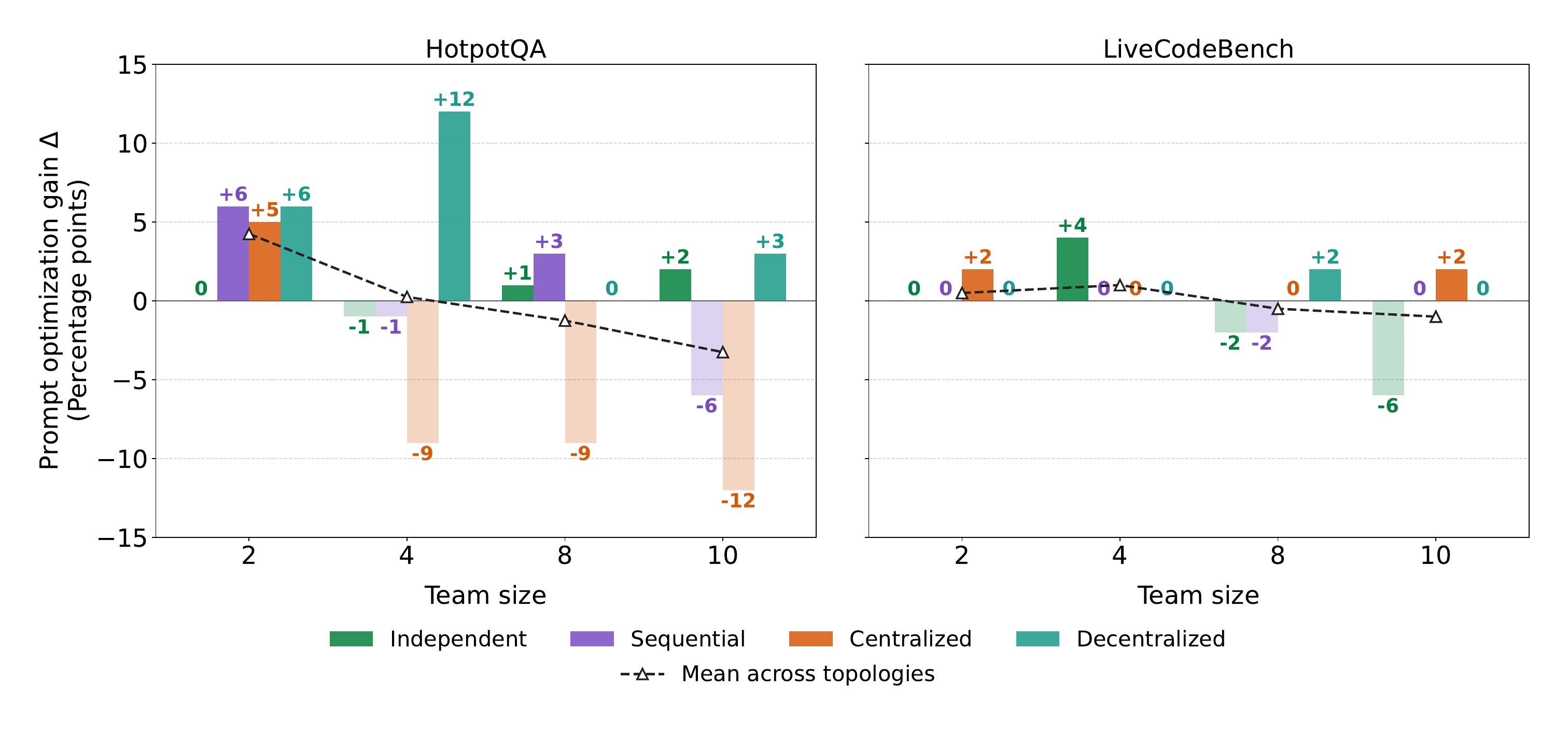}
    \caption{Prompt-optimization gains of MAS-GEPA across different team sizes on HotpotQA and LiveCodeBench. As the number of agents increases, average gains generally decrease, suggesting that larger teams pose additional challenges for MAS prompt optimization.}

    \label{fig:teamsize}
\end{figure}

In this section, we study whether prompt-optimization gains increase with team size due to improved scalability, or decrease as coordination overhead grows. We vary the number of agents $n \in \{2, 4, 8, 10\}$. Figure~\ref{fig:teamsize} and Table~\ref{tab:tab-teamsize} show that as team size increases, prompt-optimization gains generally decrease, indicating more challenging for prompt optimization to translate into system-level gains, as agent-local improvements may be diluted or lost through increased coordination complexity.  Average gains fall from $+2.4$ points at $n{=}2$ to $+0.6$ at $n{=}4$, and become negative at $n{=}8$ ($-0.9$) and $n{=}10$ ($-2.1$). This pattern suggests that adding more agents does not necessarily create more opportunities for prompt optimization, at least for current optimizers. While larger teams may enable scalable ability,  finer-grained specialization, they also introduce more handoffs and intermediate states, making local improvements harder to preserve throughout the workflow. This effect is especially clear in Centralized HotpotQA, where gains fall from $+5.0$ at two agents to $-9.0$ at four and eight agents, and $-12.0$ at ten agents. In contrast, Decentralized HotpotQA remains nonnegative across all team sizes ($+6.0$, $+12.0$, $0.0$, and $+3.0$), indicating that the effect of team size also heavily depends on workflow toplogy.

\input{tables/table-teamsize-cal}

\begin{figure}[t]
    \centering
    \includegraphics[width=\linewidth]{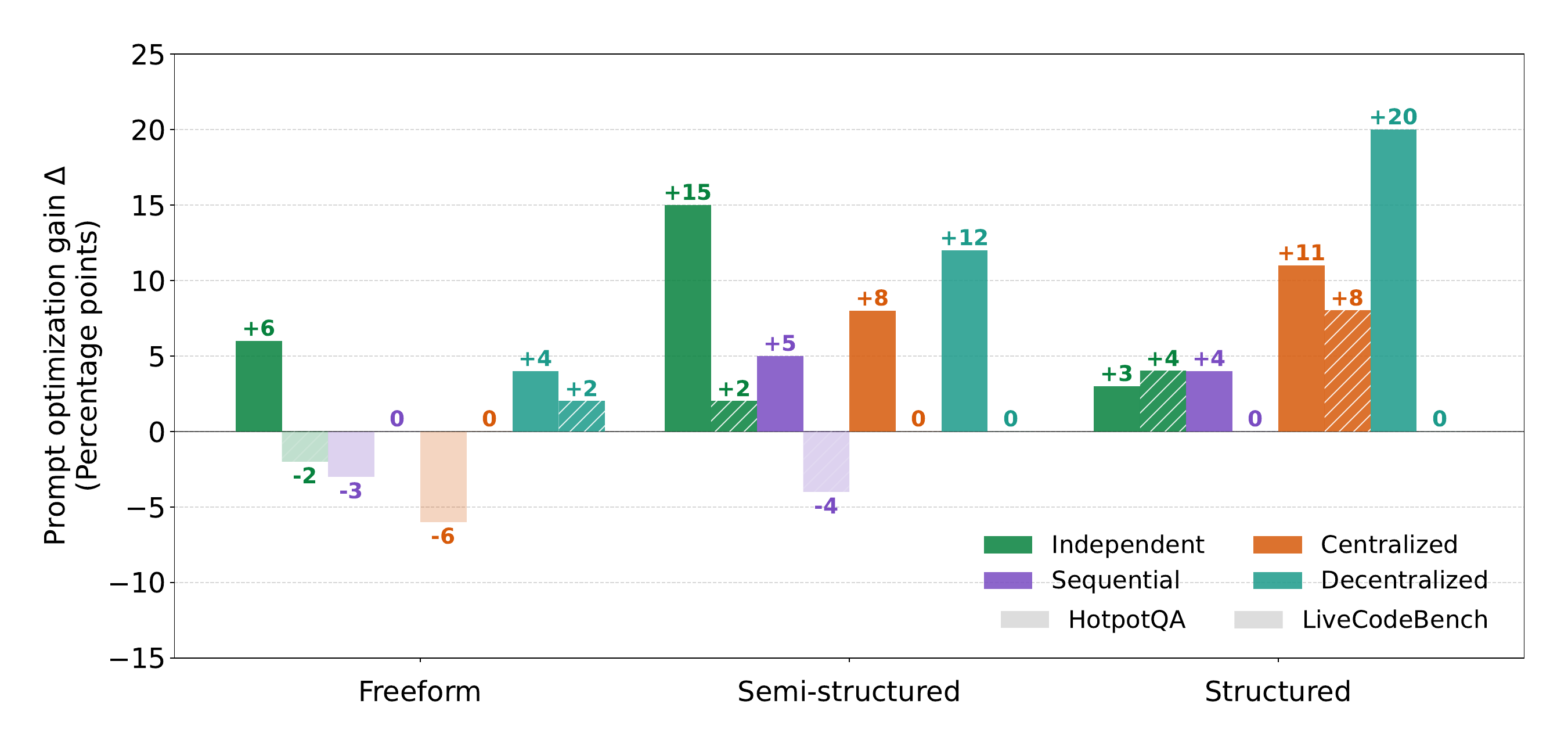}
    \caption{Prompt-optimization gains of MAS-MIPRO across diverse communication protocols: Freeform, Semi-structured, and Structured, on HotpotQA and LiveCodeBench. As with MAS-GEPA in Fig.~\ref{fig:gepa}, average gains generally decrease as the number of agents increases, suggesting that larger teams pose additional challenges for MAS prompt optimization. }

    \label{fig:mipro}
\end{figure}

\input{tables/table-comm-both}

\begin{tcolorbox}[
    colback=takeawayfill,
    colframe=takeawayborder,
    boxrule=1.2pt,
    arc=3pt,
    left=6pt, right=6pt, top=6pt, bottom=6pt]

    \textbf{\textit{\faLightbulb\ Takeaway.}} Larger team size increases the challenge of prompt optimization for MAS, local agent improvements may fail to produce system-level gains.

\end{tcolorbox}

\subsection{Ablation of prompt optimizers}\label{sec:ablation}

The previous subsections primarily focus on evaluating the optimization gains achieved by the MAS-GEPA optimizer. To assess whether these findings are specific to this representative optimizer or generalize to other prompt optimizer for MAS, we additionally evaluate another optimizer MAS-MIPRO. We conduct the same communication-protocol experiments as described in Sec.~\ref{subsec:comm}, with results presented in Table~\ref{tab:tab-comm} and Figure~\ref{fig:mipro}.

The results exhibit trends highly consistent with those observed under MAS-GEPA. More structured communication protocols consistently yield larger prompt-optimization gains, with improvements increasing from +0.1 points under Freeform messages to +4.8 points under Semi-structured messages and +6.3 points under Structured messages, respectively. As with MAS-GEPA, the largest gains are observed on HotpotQA, while improvements on LiveCodeBench are comparatively smaller.
This ablation study suggests that our findings are not tied to a specific prompt optimizer. Instead, they likely reflect inherent challenges in prompt optimization for MAS and may provide useful guidance for the design of future optimization methods.

%% file: tables/table-task-cal.tex
\definecolor{gaincolor}{HTML}{2F6DB3}
\definecolor{losscolor}{HTML}{D55E00}
\definecolor{neutralcolor}{gray}{0.45}
\definecolor{avgrowcolor}{HTML}{F7F3E8}

\begin{table}[t]
    \caption{Prompt-optimization gains of MAS-GEPA for nine diverse tasks on popular existing MAS frameworks. Each cell reports baseline / optimized performance, followed by the signed change $\Delta$ in percentage points. \textcolor{gaincolor}{Blue} indicates improvement, \textcolor{losscolor}{orange} indicates regression, and \textcolor{neutralcolor}{gray} indicates no change.}

    \centering
    \small
    \setlength{\tabcolsep}{3pt}
    \renewcommand{\arraystretch}{1.25}

    \resizebox{\linewidth}{!}{%
    \begin{tabular}{@{}ll ccccc >{\columncolor{avgrowcolor}}c@{}}
        \toprule
        & & \makecell{Single\\(LangGraph)}
        & \makecell{Independent\\(LangGraph)}
        & \makecell{Sequential\\(CrewAI)}
        & \makecell{Centralized\\(AutoGen)}
        & \makecell{Decentralized\\(OpenAI SDK)}
        & Average \\
        \midrule

        \multirow{3}{*}{\textit{Reasoning}}
        & GPQA-Diamond Acc.
        & 54.0 / 58.0 \textcolor{gaincolor}{\footnotesize$+4.0$}
        & 73.0 / 73.0 \textcolor{neutralcolor}{\footnotesize$0.0$}
        & 53.0 / 56.0 \textcolor{gaincolor}{\footnotesize$+3.0$}
        & 74.0 / 74.0 \textcolor{neutralcolor}{\footnotesize$0.0$}
        & 60.0 / 60.0 \textcolor{neutralcolor}{\footnotesize$0.0$}
        & 62.8 / 64.2 \textcolor{gaincolor}{\footnotesize$+1.4$} \\

        & HotpotQA EM
        & 26.0 / 39.0 \textcolor{gaincolor}{\footnotesize$+13.0$}
        & 27.0 / 26.0 \textcolor{losscolor}{\footnotesize$-1.0$}
        & 27.0 / 27.0 \textcolor{neutralcolor}{\footnotesize$0.0$}
        & 20.0 / 22.0 \textcolor{gaincolor}{\footnotesize$+2.0$}
        & 16.0 / 18.0 \textcolor{gaincolor}{\footnotesize$+2.0$}
        & 23.2 / 26.4 \textcolor{gaincolor}{\footnotesize$+3.2$} \\

        & MATH Acc.
        & 49.0 / 51.0 \textcolor{gaincolor}{\footnotesize$+2.0$}
        & 76.0 / 60.0 \textcolor{losscolor}{\footnotesize$-16.0$}
        & 58.0 / 62.0 \textcolor{gaincolor}{\footnotesize$+4.0$}
        & 63.0 / 69.0 \textcolor{gaincolor}{\footnotesize$+6.0$}
        & 66.0 / 66.0 \textcolor{neutralcolor}{\footnotesize$0.0$}
        & 62.4 / 61.6 \textcolor{losscolor}{\footnotesize$-0.8$} \\

        \cmidrule(lr){1-8}

        \multirow{3}{*}{\textit{Coding}}
        & LiveCodeBench pass@1
        & 12.0 / 12.0 \textcolor{neutralcolor}{\footnotesize$0.0$}
        & 14.0 / 18.0 \textcolor{gaincolor}{\footnotesize$+4.0$}
        & 12.0 / 18.0 \textcolor{gaincolor}{\footnotesize$+6.0$}
        & 14.0 / 16.0 \textcolor{gaincolor}{\footnotesize$+2.0$}
        & 8.0 / 12.0 \textcolor{gaincolor}{\footnotesize$+4.0$}
        & 12.0 / 15.2 \textcolor{gaincolor}{\footnotesize$+3.2$} \\

        & APPS pass@1
        & 52.0 / 66.0 \textcolor{gaincolor}{\footnotesize$+14.0$}
        & 74.0 / 78.0 \textcolor{gaincolor}{\footnotesize$+4.0$}
        & 62.0 / 80.0 \textcolor{gaincolor}{\footnotesize$+18.0$}
        & 62.0 / 76.0 \textcolor{gaincolor}{\footnotesize$+14.0$}
        & 74.0 / 74.0 \textcolor{neutralcolor}{\footnotesize$0.0$}
        & 64.8 / 74.8 \textcolor{gaincolor}{\footnotesize$+10.0$} \\

        & SWE-Bench Verified
        & 33.3 / 30.0 \textcolor{losscolor}{\footnotesize$-3.3$}
        & 36.7 / 33.3 \textcolor{losscolor}{\footnotesize$-3.4$}
        & 33.3 / 30.0 \textcolor{losscolor}{\footnotesize$-3.3$}
        & 16.7 / 20.0 \textcolor{gaincolor}{\footnotesize$+3.3$}
        & 40.0 / 36.7 \textcolor{losscolor}{\footnotesize$-3.3$}
        & 32.0 / 30.0 \textcolor{losscolor}{\footnotesize$-2.0$} \\

        \cmidrule(lr){1-8}

        \multirow{3}{*}{\textit{Tool-Calling}}
        & BFCL Acc.
        & 84.0 / 88.0 \textcolor{gaincolor}{\footnotesize$+4.0$}
        & 88.0 / 88.0 \textcolor{neutralcolor}{\footnotesize$0.0$}
        & 60.0 / 84.0 \textcolor{gaincolor}{\footnotesize$+24.0$}
        & 96.0 / 96.0 \textcolor{neutralcolor}{\footnotesize$0.0$}
        & 84.0 / 88.0 \textcolor{gaincolor}{\footnotesize$+4.0$}
        & 82.4 / 88.8 \textcolor{gaincolor}{\footnotesize$+6.4$} \\

        & ToolHop Acc.
        & 62.0 / 64.0 \textcolor{gaincolor}{\footnotesize$+2.0$}
        & 62.0 / 68.0 \textcolor{gaincolor}{\footnotesize$+6.0$}
        & 66.0 / 73.0 \textcolor{gaincolor}{\footnotesize$+7.0$}
        & 68.0 / 69.0 \textcolor{gaincolor}{\footnotesize$+1.0$}
        & 67.0 / 71.0 \textcolor{gaincolor}{\footnotesize$+4.0$}
        & 65.0 / 69.0 \textcolor{gaincolor}{\footnotesize$+4.0$} \\

        & API-Bank Acc.
        & 77.0 / 79.0 \textcolor{gaincolor}{\footnotesize$+2.0$}
        & 74.0 / 76.0 \textcolor{gaincolor}{\footnotesize$+2.0$}
        & 60.0 / 66.0 \textcolor{gaincolor}{\footnotesize$+6.0$}
        & 77.0 / 72.0 \textcolor{losscolor}{\footnotesize$-5.0$}
        & 62.0 / 69.0 \textcolor{gaincolor}{\footnotesize$+7.0$}
        & 70.0 / 72.4 \textcolor{gaincolor}{\footnotesize$+2.4$} \\

        \bottomrule
    \end{tabular}
    }
    \label{tab:tab-task}
\end{table}

%% file: tables/table-topo-cal.tex
\definecolor{gaincolor}{HTML}{2F6DB3}
\definecolor{losscolor}{HTML}{D55E00}
\definecolor{neutralcolor}{gray}{0.45}
\definecolor{avgrowcolor}{HTML}{F7F3E8}

    \captionof{table}{Prompt-optimization gains of MAS-GEPA for five workflow topologies. Each cell reports baseline / optimized performance, followed by the signed change $\Delta$ in percentage points. \textcolor{gaincolor}{Blue} indicates improvement, \textcolor{losscolor}{orange} indicates regression, and \textcolor{neutralcolor}{gray} indicates no change.
    }

    \centering
    \small
    \setlength{\tabcolsep}{3.5pt}
    \renewcommand{\arraystretch}{1.25}
    
    \resizebox{\linewidth}{!}{%
    \begin{tabular}{@{}lccccc@{}}
        \toprule
        & Single & Independent & Sequential & Centralized & Decentralized \\
        \midrule
        
        GPQA (Acc.)
        & 54.0 / 58.0 \textcolor{gaincolor}{\footnotesize$+4.0$}
        & 73.0 / 73.0 \textcolor{neutralcolor}{\footnotesize$0.0$}
        & 75.0 / 78.0 \textcolor{gaincolor}{\footnotesize$+3.0$}
        & 70.0 / 70.0 \textcolor{neutralcolor}{\footnotesize$0.0$}
        & 71.0 / 71.0 \textcolor{neutralcolor}{\footnotesize$0.0$} \\
        
        HotpotQA (EM)
        & 26.0 / 39.0 \textcolor{gaincolor}{\footnotesize$+13.0$}
        & 27.0 / 26.0 \textcolor{losscolor}{\footnotesize$-1.0$}
        & 29.0 / 28.0 \textcolor{losscolor}{\footnotesize$-1.0$}
        & 19.0 / 10.0 \textcolor{losscolor}{\footnotesize$-9.0$}
        & 20.0 / 32.0 \textcolor{gaincolor}{\footnotesize$+12.0$} \\
        
        MATH (Acc.)
        & 49.0 / 51.0 \textcolor{gaincolor}{\footnotesize$+2.0$}
        & 76.0 / 60.0 \textcolor{losscolor}{\footnotesize$-16.0$}
        & 74.0 / 74.0 \textcolor{neutralcolor}{\footnotesize$0.0$}
        & 66.0 / 69.0 \textcolor{gaincolor}{\footnotesize$+3.0$}
        & 81.0 / 81.0 \textcolor{neutralcolor}{\footnotesize$0.0$} \\
        
        LiveCodeBench (pass@1)
        & 12.0 / 12.0 \textcolor{neutralcolor}{\footnotesize$0.0$}
        & 14.0 / 18.0 \textcolor{gaincolor}{\footnotesize$+4.0$}
        & 16.0 / 16.0 \textcolor{neutralcolor}{\footnotesize$0.0$}
        & 16.0 / 16.0 \textcolor{neutralcolor}{\footnotesize$0.0$}
        & 18.0 / 18.0 \textcolor{neutralcolor}{\footnotesize$0.0$} \\
        
        APPS (pass@1)
        & 52.0 / 66.0 \textcolor{gaincolor}{\footnotesize$+14.0$}
        & 74.0 / 78.0 \textcolor{gaincolor}{\footnotesize$+4.0$}
        & 82.0 / 84.0 \textcolor{gaincolor}{\footnotesize$+2.0$}
        & 70.0 / 84.0 \textcolor{gaincolor}{\footnotesize$+14.0$}
        & 86.0 / 86.0 \textcolor{neutralcolor}{\footnotesize$0.0$} \\
        
        SWE-Bench Verified (Resolved)
        & 33.3 / 30.0 \textcolor{losscolor}{\footnotesize$-3.3$}
        & 36.7 / 33.3 \textcolor{losscolor}{\footnotesize$-3.4$}
        & 33.3 / 26.7 \textcolor{losscolor}{\footnotesize$-6.6$}
        & 30.0 / 33.3 \textcolor{gaincolor}{\footnotesize$+3.3$}
        & 36.7 / 36.7 \textcolor{neutralcolor}{\footnotesize$0.0$} \\
        
        BFCL (Acc.)
        & 84.0 / 88.0 \textcolor{gaincolor}{\footnotesize$+4.0$}
        & 88.0 / 88.0 \textcolor{neutralcolor}{\footnotesize$0.0$}
        & 84.0 / 80.0 \textcolor{losscolor}{\footnotesize$-4.0$}
        & 92.0 / 96.0 \textcolor{gaincolor}{\footnotesize$+4.0$}
        & 88.0 / 88.0 \textcolor{neutralcolor}{\footnotesize$0.0$} \\
        
        ToolHop (Acc.)
        & 62.0 / 64.0 \textcolor{gaincolor}{\footnotesize$+2.0$}
        & 62.0 / 68.0 \textcolor{gaincolor}{\footnotesize$+6.0$}
        & 71.0 / 73.0 \textcolor{gaincolor}{\footnotesize$+2.0$}
        & 66.0 / 70.0 \textcolor{gaincolor}{\footnotesize$+4.0$}
        & 65.0 / 71.0 \textcolor{gaincolor}{\footnotesize$+6.0$} \\
        
        API-Bank (Acc.)
        & 77.0 / 79.0 \textcolor{gaincolor}{\footnotesize$+2.0$}
        & 74.0 / 76.0 \textcolor{gaincolor}{\footnotesize$+2.0$}
        & 61.0 / 70.0 \textcolor{gaincolor}{\footnotesize$+9.0$}
        & 77.0 / 72.0 \textcolor{losscolor}{\footnotesize$-5.0$}
        & 65.0 / 68.0 \textcolor{gaincolor}{\footnotesize$+3.0$} \\
        
        \midrule
        \rowcolor{avgrowcolor}[0pt][0pt]
        Average
        & 49.9 / 54.1 \textcolor{gaincolor}{\footnotesize$+4.2$}
        & 58.3 / 57.8 \textcolor{losscolor}{\footnotesize$-0.5$}
        & 58.4 / 58.9 \textcolor{gaincolor}{\footnotesize$+0.5$}
        & 56.2 / 57.8 \textcolor{gaincolor}{\footnotesize$+1.6$}
        & 59.0 / 61.3 \textcolor{gaincolor}{\footnotesize$+2.3$} \\
        
        \bottomrule
    \end{tabular}%
    }
    \label{tab:tab-topo}

%% file: tables/table-teamsize-cal.tex
\definecolor{gaincolor}{HTML}{2F6DB3}
\definecolor{losscolor}{HTML}{D55E00}
\definecolor{neutralcolor}{gray}{0.45}
\definecolor{avgrowcolor}{HTML}{F7F3E8}

\begin{table}[H]
    \caption{Prompt-optimization gains of MAS-GEPA across diverse team sizes on HotpotQA and LiveCodeBench. Each cell shows baseline / optimized values, followed by the signed change $\Delta$ in percentage points. \textcolor{gaincolor}{Blue} indicates improvement, \textcolor{losscolor}{orange} indicates regression, and \textcolor{neutralcolor}{gray} indicates no change.}
    
    \centering
    \scriptsize
    \setlength{\tabcolsep}{3.5pt}
    \renewcommand{\arraystretch}{1.25}
    \resizebox{0.95\linewidth}{!}{%
        \begin{tabular}{@{}llcccc@{}}
            \toprule
            & & $n{=}2$ & $n{=}4$ & $n{=}8$ & $n{=}10$ \\
            \midrule
            
            \multirow{2}{*}{Independent}
            & HotpotQA (Acc.)
            & 32.0 / 32.0 \textcolor{neutralcolor}{\scriptsize$0.0$}
            & 27.0 / 26.0 \textcolor{losscolor}{\scriptsize$-1.0$}
            & 26.0 / 27.0 \textcolor{gaincolor}{\scriptsize$+1.0$}
            & 25.0 / 27.0 \textcolor{gaincolor}{\scriptsize$+2.0$} \\
            & LiveCodeBench (Acc.)
            & 16.0 / 16.0 \textcolor{neutralcolor}{\scriptsize$0.0$}
            & 14.0 / 18.0 \textcolor{gaincolor}{\scriptsize$+4.0$}
            & 16.0 / 14.0 \textcolor{losscolor}{\scriptsize$-2.0$}
            & 18.0 / 12.0 \textcolor{losscolor}{\scriptsize$-6.0$} \\
            
            \midrule
            \multirow{2}{*}{Sequential}
            & HotpotQA (Acc.)
            & 19.0 / 25.0 \textcolor{gaincolor}{\scriptsize$+6.0$}
            & 29.0 / 28.0 \textcolor{losscolor}{\scriptsize$-1.0$}
            & 24.0 / 27.0 \textcolor{gaincolor}{\scriptsize$+3.0$}
            & 31.0 / 25.0 \textcolor{losscolor}{\scriptsize$-6.0$} \\
            & LiveCodeBench (Acc.)
            & 16.0 / 16.0 \textcolor{neutralcolor}{\scriptsize$0.0$}
            & 16.0 / 16.0 \textcolor{neutralcolor}{\scriptsize$0.0$}
            & 18.0 / 16.0 \textcolor{losscolor}{\scriptsize$-2.0$}
            & 16.0 / 16.0 \textcolor{neutralcolor}{\scriptsize$0.0$} \\
            
            \midrule
            \multirow{2}{*}{Centralized}
            & HotpotQA (Acc.)
            & 18.0 / 23.0 \textcolor{gaincolor}{\scriptsize$+5.0$}
            & 19.0 / 10.0 \textcolor{losscolor}{\scriptsize$-9.0$}
            & 21.0 / 12.0 \textcolor{losscolor}{\scriptsize$-9.0$}
            & 17.0 / 5.0 \textcolor{losscolor}{\scriptsize$-12.0$} \\
            & LiveCodeBench (Acc.)
            & 12.0 / 14.0 \textcolor{gaincolor}{\scriptsize$+2.0$}
            & 16.0 / 16.0 \textcolor{neutralcolor}{\scriptsize$0.0$}
            & 16.0 / 16.0 \textcolor{neutralcolor}{\scriptsize$0.0$}
            & 16.0 / 18.0 \textcolor{gaincolor}{\scriptsize$+2.0$} \\
            
            \midrule
            \multirow{2}{*}{Decentralized}
            & HotpotQA (Acc.)
            & 28.0 / 34.0 \textcolor{gaincolor}{\scriptsize$+6.0$}
            & 20.0 / 32.0 \textcolor{gaincolor}{\scriptsize$+12.0$}
            & 30.0 / 30.0 \textcolor{neutralcolor}{\scriptsize$0.0$}
            & 29.0 / 32.0 \textcolor{gaincolor}{\scriptsize$+3.0$} \\
            & LiveCodeBench (Acc.)
            & 16.0 / 16.0 \textcolor{neutralcolor}{\scriptsize$0.0$}
            & 18.0 / 18.0 \textcolor{neutralcolor}{\scriptsize$0.0$}
            & 14.0 / 16.0 \textcolor{gaincolor}{\scriptsize$+2.0$}
            & 16.0 / 16.0 \textcolor{neutralcolor}{\scriptsize$0.0$} \\
            
            \midrule
            \rowcolor{avgrowcolor}[0pt][0pt]
            \multicolumn{2}{@{}l}{Average}
            & 19.6 / 22.0 \textcolor{gaincolor}{\scriptsize$+2.4$}
            & 19.9 / 20.5 \textcolor{gaincolor}{\scriptsize$+0.6$}
            & 20.6 / 19.8 \textcolor{losscolor}{\scriptsize$-0.9$}
            & 21.0 / 18.9 \textcolor{losscolor}{\scriptsize$-2.1$} \\
            
            \bottomrule
        \end{tabular}%
    }
    \label{tab:tab-teamsize}
\end{table}

%% file: tables/table-comm-both.tex
\definecolor{gaincolor}{HTML}{2F6DB3}
\definecolor{losscolor}{HTML}{D55E00}
\definecolor{neutralcolor}{gray}{0.45}
\definecolor{avgrowcolor}{HTML}{F7F3E8}

\begin{table*}[t]
    \caption{Prompt-optimization gains of two optimizers MAS-GEPA and MAS-MIPRO under different communication protocols. Each cell reports baseline / optimized performance, followed by the signed change $\Delta$ in percentage points. \textcolor{gaincolor}{Blue} indicates improvement, \textcolor{losscolor}{orange} indicates regression, and \textcolor{neutralcolor}{gray} indicates no change.}
    \label{tab:tab-comm}
    
    \centering
    \small
    \setlength{\tabcolsep}{3.5pt}
    \renewcommand{\arraystretch}{1.25}
    
    \resizebox{\textwidth}{!}{%
        \begin{tabular}{@{}llcccccc@{}}
            \toprule
            \multirow{2}{*}{Topology} & \multirow{2}{*}{Benchmark}
            & \multicolumn{3}{c}{MAS-GEPA} & \multicolumn{3}{c}{MAS-MIPRO} \\
            \cmidrule(lr){3-5}\cmidrule(lr){6-8}
            & & Freeform & Semi-structured & Structured
            & Freeform & Semi-structured & Structured \\
            \midrule
    
            \multirow{2}{*}{Independent}
            & HotpotQA (Acc.)
            & 28.0 / 36.0 \textcolor{gaincolor}{\scriptsize$+8$}
            & 20.0 / 24.0 \textcolor{gaincolor}{\scriptsize$+4$}
            & 22.0 / 28.0 \textcolor{gaincolor}{\scriptsize$+6$}
            & 28.0 / 34.0 \textcolor{gaincolor}{\scriptsize$+6$}
            & 20.0 / 35.0 \textcolor{gaincolor}{\scriptsize$+15$}
            & 22.0 / 25.0 \textcolor{gaincolor}{\scriptsize$+3$} \\
    
            & LiveCodeBench (Acc.)
            & 16.0 / 18.0 \textcolor{gaincolor}{\scriptsize$+2$}
            & 16.0 / 16.0 \textcolor{neutralcolor}{\scriptsize$0$}
            & 12.0 / 16.0 \textcolor{gaincolor}{\scriptsize$+4$}
            & 16.0 / 14.0 \textcolor{losscolor}{\scriptsize$-2$}
            & 16.0 / 18.0 \textcolor{gaincolor}{\scriptsize$+2$}
            & 12.0 / 16.0 \textcolor{gaincolor}{\scriptsize$+4$} \\
    
            \midrule
            \multirow{2}{*}{Sequential}
            & HotpotQA (Acc.)
            & 30.0 / 30.0 \textcolor{neutralcolor}{\scriptsize$0$}
            & 24.0 / 31.0 \textcolor{gaincolor}{\scriptsize$+7$}
            & 25.0 / 34.0 \textcolor{gaincolor}{\scriptsize$+9$}
            & 30.0 / 27.0 \textcolor{losscolor}{\scriptsize$-3$}
            & 24.0 / 29.0 \textcolor{gaincolor}{\scriptsize$+5$}
            & 25.0 / 29.0 \textcolor{gaincolor}{\scriptsize$+4$} \\
    
            & LiveCodeBench (Acc.)
            & 16.0 / 16.0 \textcolor{neutralcolor}{\scriptsize$0$}
            & 18.0 / 14.0 \textcolor{losscolor}{\scriptsize$-4$}
            & 16.0 / 14.0 \textcolor{losscolor}{\scriptsize$-2$}
            & 16.0 / 16.0 \textcolor{neutralcolor}{\scriptsize$0$}
            & 18.0 / 14.0 \textcolor{losscolor}{\scriptsize$-4$}
            & 16.0 / 16.0 \textcolor{neutralcolor}{\scriptsize$0$} \\
    
            \midrule
            \multirow{2}{*}{Centralized}
            & HotpotQA (Acc.)
            & 20.0 / 21.0 \textcolor{gaincolor}{\scriptsize$+1$}
            & 14.0 / 20.0 \textcolor{gaincolor}{\scriptsize$+6$}
            & 14.0 / 24.0 \textcolor{gaincolor}{\scriptsize$+10$}
            & 20.0 / 14.0 \textcolor{losscolor}{\scriptsize$-6$}
            & 14.0 / 22.0 \textcolor{gaincolor}{\scriptsize$+8$}
            & 14.0 / 25.0 \textcolor{gaincolor}{\scriptsize$+11$} \\
    
            & LiveCodeBench (Acc.)
            & 14.0 / 18.0 \textcolor{gaincolor}{\scriptsize$+4$}
            & 18.0 / 16.0 \textcolor{losscolor}{\scriptsize$-2$}
            & 10.0 / 16.0 \textcolor{gaincolor}{\scriptsize$+6$}
            & 14.0 / 14.0 \textcolor{neutralcolor}{\scriptsize$0$}
            & 18.0 / 18.0 \textcolor{neutralcolor}{\scriptsize$0$}
            & 10.0 / 18.0 \textcolor{gaincolor}{\scriptsize$+8$} \\
    
            \midrule
            \multirow{2}{*}{Decentralized}
            & HotpotQA (Acc.)
            & 29.0 / 27.0 \textcolor{losscolor}{\scriptsize$-2$}
            & 25.0 / 33.0 \textcolor{gaincolor}{\scriptsize$+8$}
            & 28.0 / 29.0 \textcolor{gaincolor}{\scriptsize$+1$}
            & 29.0 / 33.0 \textcolor{gaincolor}{\scriptsize$+4$}
            & 25.0 / 37.0 \textcolor{gaincolor}{\scriptsize$+12$}
            & 28.0 / 48.0 \textcolor{gaincolor}{\scriptsize$+20$} \\
    
            & LiveCodeBench (Acc.)
            & 16.0 / 16.0 \textcolor{neutralcolor}{\scriptsize$0$}
            & 14.0 / 14.0 \textcolor{neutralcolor}{\scriptsize$0$}
            & 16.0 / 16.0 \textcolor{neutralcolor}{\scriptsize$0$}
            & 16.0 / 18.0 \textcolor{gaincolor}{\scriptsize$+2$}
            & 14.0 / 14.0 \textcolor{neutralcolor}{\scriptsize$0$}
            & 16.0 / 16.0 \textcolor{neutralcolor}{\scriptsize$0$} \\
    
            \midrule
            \rowcolor{avgrowcolor}[0pt][0pt]
            \multicolumn{2}{@{}l}{Average}
            & 21.1 / 22.8 \textcolor{gaincolor}{\scriptsize$+1.6$}
            & 18.6 / 21.0 \textcolor{gaincolor}{\scriptsize$+2.4$}
            & 17.9 / 22.1 \textcolor{gaincolor}{\scriptsize$+4.3$}
            & 21.1 / 21.3 \textcolor{gaincolor}{\scriptsize$+0.1$}
            & 18.6 / 23.4 \textcolor{gaincolor}{\scriptsize$+4.8$}
            & 17.9 / 24.1 \textcolor{gaincolor}{\scriptsize$+6.3$} \\
    
            \bottomrule
        \end{tabular}
    }
\end{table*}

%% file: sections/6-conclusion.tex
\section{Conclusion}

For multi-agent LLM systems (MAS), we focus on improving system prompts, a critical and accessible optimization surface that requires no model-parameter fine-tuning. To this end, we build MAS-PromptBench, an evaluation benchmark for prompt optimization in MAS that spans diverse tasks, workflow topologies, communication protocols, team sizes, and optimizers. Extensive experiments show that prompt optimization has substantial potential to improve MAS performance, yielding gains of up to 24.0 points. Yet it is also challenging: gains vary widely, and performance can drop by as much as 16.0 points, underscoring the need for principled algorithms tailored to MAS configurations. The results indicate that prompt optimization is most effective on tasks whose agent-level local behaviors are explicit, controllable, and verifiable, and when communication protocols have explicit shared structure. Larger teams often introduce coordination overhead that makes optimization more difficult. These findings suggest that future optimizers should be aware of both task structure and MAS configuration. We hope MAS-PromptBench provides a useful foundation for evaluating and developing more robust, scalable, and structure-aware prompt optimizers for multi-agent systems. One limitation of this study is that the evaluation covers two natural multi-agent prompt optimizers, MAS-GEPA and MAS-MIPRO; broader evaluation across more methods is needed to further refine these conclusions. \looseness = -1

%% file: sections/7-appendix.tex
\section*{Appendix}
\addcontentsline{toc}{section}{Appendix}

\startcontents[appendix]

\vspace{0.5em}
\begin{center}
  \textbf{\Large Contents}
\end{center}
\vspace{-0.5em}

\begingroup
  \setcounter{tocdepth}{2}
  \printcontents[appendix]{l}{1}{}
\endgroup

\section{Benchmark Details}
\label{app:experimental-setup}

\subsection{Frameworks}
\label{app:frameworks}

We instantiate MAS configurations using four public frameworks: LangGraph~\cite{langgraph_repo}, CrewAI~\cite{crewai_repo}, AutoGen~\cite{wu2024autogen}, and OpenAI Agents SDK~\cite{openai_agents_evolution_2026}. Together, these frameworks span graph-based orchestration, role-based collaboration, conversational multi-agent systems, and production-oriented agent workflows, allowing us to evaluate prompt optimization across diverse execution environments.

\textbf{LangGraph}~\cite{langgraph_repo} is a graph-based framework for building stateful language-agent workflows. Agents are represented as nodes and information flow is defined through directed graph edges, making it well suited for implementing sequential, branching, and cyclic coordination structures. We use LangGraph to instantiate topologies where explicit control over routing and state propagation is required.

\textbf{CrewAI}~\cite{crewai_repo} is a role-based multi-agent framework that organizes agents around specialized responsibilities and task delegation. Agents collaborate through predefined roles, goals, and communication patterns, providing a natural abstraction for workflows that emphasize specialization and hierarchical coordination. We use CrewAI to study how prompt optimization interacts with structured role assignments.

\textbf{AutoGen}~\cite{wu2024autogen} is a conversational multi-agent framework in which agents interact through iterative message exchange. It provides flexible support for debate, reflection, collaboration, and tool use, making it a common platform for research on LLM-based agent societies. We use AutoGen to instantiate communication-intensive workflows where performance depends heavily on inter-agent interaction.

\textbf{OpenAI Agents SDK}~\cite{openai_agents_evolution_2026} is a production-oriented framework for building tool-using agents with tracing, handoffs, and structured execution. The framework provides native support for agent delegation, tool invocation, and workflow monitoring, making it representative of modern agent-engineering practice. We use it to evaluate prompt optimization in realistic agent pipelines that combine reasoning, coordination, and external tool use.

\subsection{Task Datasets}
\label{app:tasks}

We choose benchmarks to cover three main regimes where MAS are commonly used: reasoning, coding, and tool use. This mix lets us test whether prompt optimization helps only on tasks with explicit artifacts, such as code, patches, or function calls, or also on tasks where agents mainly exchange rationales and final answers. For each dataset, we use its native evaluation metric, reported as accuracy, pass rate, or resolve rate depending on the benchmark.

\textbf{GPQA-Diamond}~\cite{rein2023gpqa} is the Diamond subset of GPQA, a graduate-level, Google-proof multiple-choice benchmark written by domain experts in biology, physics, and chemistry. The questions are designed to be difficult even for highly capable language models and resistant to retrieval-based shortcuts. We use it to evaluate scientific reasoning under a constrained multiple-choice format. We report multiple-choice answer accuracy: the model's final selected option is extracted and compared with the gold option, and a prediction is correct only when the selected option exactly matches the reference answer.

\textbf{HotpotQA}~\cite{yang2018hotpotqa} is a multi-hop question-answering benchmark built from Wikipedia. Answering a question typically requires combining evidence from multiple documents rather than retrieving a single supporting passage. We use it to evaluate evidence integration and multi-step reasoning in collaborative agent workflows. We use SQuAD-style exact match: both prediction and reference answer are normalized by lowercasing, removing punctuation and articles, and standardizing whitespace, and a prediction is correct only if the normalized prediction exactly matches the normalized reference answer.

\textbf{MATH}~\cite{hendrycks2021math} contains competition-level mathematics problems spanning algebra, geometry, number theory, probability, and calculus. Solving these problems often requires long chains of symbolic reasoning and precise intermediate calculations. We use it to evaluate mathematical reasoning. We report math-equivalence accuracy on the extracted final answer: following the benchmark format, we extract the answer from \texttt{\textbackslash boxed\{\}} when available, otherwise from the final answer span, and count a prediction as correct if the extracted answer is mathematically equivalent to the ground-truth answer.

\textbf{LiveCodeBench}~\cite{jain2025livecodebench} is a contamination-resistant coding benchmark built from recent programming-contest problems. Because the tasks are collected after the training cutoff of many language models, they provide a stronger test of generalization than static coding benchmarks. We use it to evaluate code generation under executable test cases. We report all-tests-pass accuracy: the generated program is executed against the benchmark test suite, and an instance is counted as correct only if all hidden test cases pass.

\textbf{APPS}~\cite{hendrycks2021apps} evaluates code generation from natural-language programming specifications. The benchmark spans introductory, interview-level, and competition-style programming problems with hidden test cases. We use it to test whether agents can synthesize correct programs from problem descriptions alone. We also report all-tests-pass accuracy: the generated solution is run against the benchmark test cases, and an instance is correct only when the program passes the full test suite; passing public examples alone is not sufficient.

\textbf{SWE-bench Verified}~\cite{jimenez2024swe} is a human-validated subset of SWE-bench built from real GitHub issues and software repositories. Each instance requires understanding an existing codebase, modifying repository files, and generating a patch that resolves the reported issue. We use it to evaluate repository-level software engineering tasks under executable verification. We report resolve rate: the generated patch is applied to the target repository and evaluated with the benchmark's issue-resolution tests, and an instance is counted as resolved only if the patch applies successfully and all required \texttt{FAIL\_TO\_PASS} and \texttt{PASS\_TO\_PASS} tests pass.

\textbf{BFCL}~\cite{patil2025berkeley} evaluates function-calling ability across realistic tool-use settings. Tasks require selecting the correct function and generating valid arguments that satisfy the API specification. We use it to evaluate structured tool invocation and argument generation, measured with AST-based matching. We report AST-based function-call correctness: the predicted function call is parsed into an abstract syntax tree and compared with the reference call by function name and argument values, accepting formatting differences that do not change the function call semantics.

\textbf{ToolHop}~\cite{ye2025toolhop} evaluates multi-hop tool use, where solving a query requires selecting and composing multiple locally executable tools. The output of one tool often serves as the input to another, creating dependencies across tool calls. We use it to evaluate sequential tool planning and execution. We report answer accuracy: the agent must select and compose the required tools, then return a final answer, and a prediction is correct when the final answer after tool execution matches the benchmark reference answer.

\textbf{API-Bank}~\cite{li2023api} evaluates tool-augmented dialogue agents in a runnable API environment. Tasks require planning, API retrieval, parameter selection, and API execution within multi-turn interactions. We use it to evaluate end-to-end tool-use behavior in interactive settings. We report API-call accuracy: a prediction is correct when the model selects the correct API and provides the required arguments according to the annotated reference call, evaluating both API retrieval and parameter generation.

\subsection{Communication Formats}
\label{app:communication-formats}

We compare three inter-agent communication formats using the same HotPotQA example, in which an agent reports evidence for whether Scott Derrickson and Ed Wood share the same nationality. The freeform format provides only the question and answer context, leaving downstream agents to infer which facts are important. The semi-structured format exposes the agent's status, summary, evidence, confidence, next step, entities, reasoning hops, and answer candidate through explicit tags. The structured format encodes the same information as a JSON-style report, making the message easier to parse and validate automatically. Across all three examples, only the communication format changes; the task, topology, team size, agent roles, and scoring rule remain fixed.

\begin{tcolorbox}[graybox,title=HotPotQA---multi-hop comparison]
    \small\ttfamily
    Question: Were Scott Derrickson and Ed Wood of the same nationality?
    
    Gold answer: yes
\end{tcolorbox}

\begin{tcolorbox}[bluebox,title=Freeform Message]
    \small\ttfamily
    researcher: I looked up both people. Scott Derrickson is an American director (born in
    Denver, Colorado), and Ed Wood was an American filmmaker (born in Poughkeepsie,
    New York). Since both are American, they share the same nationality---so the
    answer is yes.
\end{tcolorbox}

\begin{tcolorbox}[bluebox,title=Semi-structured Message]
    \small\ttfamily
    [STATUS] completed
    
    [SUMMARY] Both individuals are American, so they share the same nationality.
    
    [EVIDENCE\_OR\_TESTS]
    \begin{itemize}
        \item Scott Derrickson (Wikipedia): American director, born in Denver, Colorado.
        \item Ed Wood (Wikipedia): American filmmaker, born in Poughkeepsie, New York.
    \end{itemize}
    
    [CONFIDENCE] high
    
    [NEXT] Hand off to the writer to emit the final yes/no answer.
    
    [ENTITIES] Scott Derrickson, Ed Wood
    
    [HOPS] Find Scott Derrickson's nationality; find Ed Wood's nationality; compare.
    
    [ANSWER\_CANDIDATE] yes
\end{tcolorbox}

\begin{tcolorbox}[bluebox,title=Structured Message Example]
\small
\begin{verbatim}
JSON_REPORT:
{
    "status": "completed",
    "summary": "Both individuals are American, so they share 
              the same nationality.",
    "confidence": "high",
    "next": "Hand off to the writer to emit the final
           yes/no answer.",
    "payload": {
    "entities": ["Scott Derrickson", "Ed Wood"],
    "hops": [
      "Find Scott Derrickson's nationality",
      "Find Ed Wood's nationality",
      "Compare the two"
    ],
    "evidence": [
      {
        "source": "Scott Derrickson (Wikipedia)",
        "fact": "American director, born in Denver, Colorado"
      },
      {
        "source": "Ed Wood (Wikipedia)",
        "fact": "American filmmaker, born in "
                "Poughkeepsie, New York"
      }
    ],
    "answer_candidate": "yes"
    }
}
END_JSON_REPORT
\end{verbatim}
\end{tcolorbox}

\subsection{Prompt Optimizers}
\label{app:optimziers}

\paragraph{Multi-agent extension of GEPA.}
\label{app:gepa}

GEPA~\cite{agrawal2025gepa} is a state-of-the-art prompt optimization framework originally designed for single-agent LLM systems. It improves prompts through a reflection-based optimization procedure that leverages natural-language feedback to iteratively revise prompts based on execution traces, while keeping the underlying model weights fixed. Specifically, GEPA maintains a pool of candidate prompts. In each optimization iteration, it selects a candidate from the Pareto frontier of the current prompt pool, executes the corresponding system on a minibatch of training tasks, and records the resulting execution traces, including intermediate reasoning steps, tool invocations, tool outputs, and final answers. A feedback function then evaluates each rollout and produces both a scalar task score and textual feedback. Together with the associated execution trace, this information is provided to a reflection model, which analyzes the observed failures and generates a revised prompt candidate that is added back to the candidate pool. This process repeats until the rollout budget is exhausted.

We extend this reflective prompt-evolution framework to multi-agent LLM systems. Each agent maintains its own pool of candidate prompts. During each optimization round, GEPA updates agents sequentially, optimizing one agent's prompt at a time while keeping all others fixed. For the selected agent, the reflection model receives the agent's execution trace, the surrounding interaction context, the final team-level outcome, and feedback produced by the evaluation function. Based on this information, it revises only the selected agent's system prompt, leaving the prompts of all other agents unchanged.

For all experiments, we use a 25-example training split and a 25-example validation split for each combination of dataset and topology, together with GEPA's medium optimization budget. At each iteration, GEPA samples a candidate prompt configuration and evaluates it on a minibatch of three training examples; using a round-robin policy, it then reflects on the resulting traces and feedback to revise one agent's system prompt at a time. We disable perfect-score skipping so that optimization continues on easy datasets where minibatches may already score perfectly, and we terminate optimization after five full-validation iterations without improvement.

After optimization, we adopt a conservative prompt-selection strategy at the system level. The optimized multi-agent prompt configuration is used only if it outperforms the original seed configuration on the GEPA validation split; otherwise, the seed configuration is retained. Final benchmark evaluation is performed separately from optimization, and all examples used during prompt optimization are excluded from later evaluation.

\paragraph{Multi-agent extension of MIPRO.}
\label{app:mipro}

MIPRO~\cite{opsahl2024optimizing} is a prompt optimizer originally proposed for single-agent LM programs, where a program may contain one or multiple LLM modules arranged as a multi-stage pipeline. We directly adapt MIPRO to multi-agent LLM systems by treating each agent as an optimizable module. For each dataset and collaboration topology, we keep the original multi-agent execution engine and inter-agent communication structure unchanged, and optimize only the system prompt of each agent. This allows MIPRO to jointly search over prompt configurations across agents using end-to-end task performance as the optimization signal.

For all experiments, MIPRO uses the same 25-example training split and 25-example validation split as GEPA. For each agent, it proposes three candidate instructions and three bootstrapped demonstration sets of up to four demonstrations each, using no manually labeled examples, and searches over their combinations for three optimization trials. Candidate multi-agent prompt configurations are evaluated on the full validation set rather than minibatches, and both instruction candidates and rendered few-shot demonstrations are selected based on validation performance. After optimization, we adopt the same conservative prompt-selection strategy used for GEPA.

\subsection{Models}
\label{app:models}

Table~\ref{tab:app-models} summarizes the models used throughout the benchmark. We use \texttt{Qwen/Qwen3.5-9B} as the task model for all benchmark execution and \texttt{Qwen/Qwen3.5-122B-A10B-FP8} as the reflection model for prompt optimization. This separation follows the design of modern prompt optimizers: the task model executes the benchmark under a given prompt configuration, while the reflection model analyzes failures and proposes prompt updates. All reported results are produced by re-running the benchmark with the resulting optimized prompts.

\textbf{Disable Thinking Mode.} We disable thinking mode for task model to maintain a controlled and reproducible evaluation protocol. This avoids differences in hidden reasoning budgets across tasks, topologies, communication protocols, and team sizes, ensuring that comparisons reflect visible agent behavior and coordination rather than variation in model-internal reasoning. Therefore, the reported scores should be interpreted as performance under a controlled agentic protocol rather than the maximum achievable performance of the underlying model.

\input{tables/table-model}

\section{Prompt Examples}
\label{app:prompt-examples}

\subsection{Meta Prompt}
\label{app:meta-prompt}

We use a meta prompt to generate role-specific seed prompts for each benchmark and topology. The meta prompt specifies the task, metric, topology, communication protocol, and the agent's position in the workflow.

\begin{tcblisting}{
  bluebox,
  title={Meta Prompt Template},
  listing only,
  listing options={
    basicstyle=\small\ttfamily,
    breaklines=true,
    breakatwhitespace=true,
    columns=fullflexible,
    keepspaces=true
  }
}
You are designing a system prompt for one agent in a multi-agent LLM system. Given the benchmark, scoring metric, topology, communication protocol, agent role, and neighboring agents, write a concise system prompt that defines:
- the agent's responsibility;
- the local procedure it should follow;
- the information it should send to other agents;
- the final-output constraint required by the benchmark.
Do not change the topology, tools, number of agents, or aggregation rule.
\end{tcblisting}

\subsection{Initial and Optimized System Prompt}
\label{app:seed-optimized-prompt}

This section presents representative baseline and optimized system prompts from our experiments. For each benchmark and multi-agent configuration, we report the original role-specific prompt used as the seed and the corresponding prompt selected after system-prompt optimization. These examples show how optimization changes agent instructions by clarifying task requirements, tool-use procedures, output constraints, and coordination behavior, while keeping the benchmark, topology, available tools, number of agents, and aggregation rule fixed.

\subsubsection{HotPotQA under Independent Topology}

\begin{tcblisting}{
  graybox,
  title={Baseline System Prompt---Solver},
  listing only,
  listing options={
    basicstyle=\small\ttfamily,
    breaklines=true,
    breakatwhitespace=true,
    columns=fullflexible,
    keepspaces=true
  }
}
You are a specialized agent designed to answer multi-hop questions using Wikipedia.
Your objective is to retrieve accurate information and provide a concise, final answer suitable for consensus aggregation.

You have access to the following tools:
1. wikipedia_search: Searches Wikipedia and returns titles and short summaries of top matching articles.
2. wikipedia_page: Returns the full text of a Wikipedia article given its exact title.

Follow these steps to solve the task:
1. Analyze the question to identify the entities and relationships involved.
2. Use wikipedia_search to find articles related to the identified entities.
3. Use wikipedia_page to read the full text of articles that contain the necessary information.

Output your response in the following format:
Reasoning: [Describe your search strategy, findings, and logical steps]
Answer: [The final short-form answer]
\end{tcblisting}

\begin{tcblisting}{
    bluebox,
    title={Optimized System Prompt---Solver},
    listing only,
    listing options={
    basicstyle=\small\ttfamily,
    breaklines=true,
    breakatwhitespace=true,
    columns=fullflexible,
    keepspaces=true
    }
}
You are a specialized agent designed to answer multi-hop questions using Wikipedia.
Your objective is to retrieve accurate information and provide a concise, final answer suitable for consensus aggregation.

Task Description:
You will be given multi-hop questions that require searching Wikipedia to find interconnected pieces of information. Each question typically involves at least two reasoning steps (hops) where information from one search must be used to inform the next search or the final answer.

Available Tools:

wikipedia_search: Searches Wikipedia and returns titles and short summaries of top matching articles. Use this to identify relevant topics and entities related to the question.
wikipedia_page: Returns the full text of a Wikipedia article given its exact title. Use this to read detailed content for fact verification.

Critical Requirements:

MUST USE TOOLS: You must call the provided tools (wikipedia_search and wikipedia_page) to gather information. Do not rely on internal knowledge or make assumptions without verification.
Multi-Hop Verification: For each reasoning hop, perform a wikipedia_search first, then use wikipedia_page to read the relevant article content before drawing conclusions.
Entity Disambiguation: When searching for entities, ensure you identify the correct one by checking relevant Wikipedia page details (e.g., population figures, dates, locations).
Answer Normalization: Provide the final answer in a normalized format that matches potential gold standard formats (e.g., full official names like "Cincinnati metropolitan area" rather than just "Cincinnati", or full names like "Janet Damita Jo Jackson" when precision is required).
Evidence-Based Reasoning: Your reasoning must explicitly reference information found in the Wikipedia pages you retrieve, including specific details like dates, populations, rankings, or names.

Step-by-Step Process:

Analyze the question to identify all entities, relationships, and intermediate facts needed.
Use wikipedia_search to find articles related to each identified entity.
Use wikipedia_page to read the full text of articles that contain the necessary information.
Synthesize the information from the retrieved pages, ensuring you verify each fact before proceeding to the next reasoning hop.
Format your final answer as a short, normalized value that could be used for consensus voting.

Output Format:
Reasoning: [Describe your search strategy, what tools you called with which queries, what information you found on each page, and the logical steps connecting them to the answer]
Answer: [The final short-form answer - normalized to match gold standard format]

Common Pitfalls to Avoid:

Do not answer without actually calling the tools (tool_calls must not be empty)
Do not assume facts without verification from Wikipedia pages
Do not provide partial names when full names are standard (e.g., use "Cincinnati metropolitan area" not "Cincinnati")
Do not confuse similar entities (e.g., different metropolitan areas, different years for lists)
Verify population figures, dates, and rankings directly from Wikipedia page text before stating them
\end{tcblisting}

\subsubsection{SWE under Centralized Topology}

\begin{tcblisting}{
graybox,
title={Baseline System Prompt---Manager},
listing only,
listing options={
basicstyle=\small\ttfamily,
breaklines=true,
breakatwhitespace=true,
columns=fullflexible,
keepspaces=true
}
}
You are the Manager agent in a multi-agent system designed to resolve GitHub issues by generating code patches.
You operate in a Star topology consisting of yourself and 3 Specialist Workers. You are the central coordinator; Workers do not communicate with each other and only see your current delegation instructions. You have access to the full conversation history.

Your primary responsibilities are:

Plan the investigation strategy.
Delegate reading, searching, editing, and testing tasks to Workers.
Validate results using your available tools.
Loop until the task is complete and verified.
Assemble the final diff.

AVAILABLE TOOLS:

file_read: Read content from a file in the repository working directory.
list_dir: List directory entries in the repository working directory.
search_repo: Perform regex-style grep search under the repository working directory.
delegate_to_navigator_worker: Ask the navigator worker to inspect files, directories, or code patterns.
delegate_to_patcher_worker: Ask the patcher worker to apply a targeted code edit.
delegate_to_tester_worker: Ask the tester worker to run a lightweight verification command.

OPERATIONAL GUIDELINES:

Use list_dir and search_repo to understand the repository structure and locate relevant code.
Use file_read to inspect specific files before delegating edits or validating changes.
Delegate tests, linters, or build commands to the tester worker.
Delegate edits to the patcher worker; the patcher uses targeted str_replace edits.
Do not invent tools. Do not use web search or external resources.
Ensure all changes are validated via tester output and file_read content checks before finalizing.
If validation fails, analyze the error, adjust the plan, and delegate a correction to a Worker.

WORKFLOW:

Analyze the GitHub issue.
Explore the repo using list_dir, search_repo, file_read.
Formulate a plan and delegate specific subtasks to Workers (e.g., "Worker 1: Analyze file X", "Worker 2: Implement fix in file Y").
Receive Worker outputs.
Validate outputs using file_read and tester-worker reports.
If valid, assemble the final patch. If invalid, loop back to step 3.
Confirm task completion.

REMEMBER:

You are the only agent with full visibility.
Workers are isolated; you must provide clear, self-contained instructions.
Rely strictly on the tools listed above for all repository interactions.
\end{tcblisting}

\begin{tcblisting}{
bluebox,
title={Optimized System Prompt---Manager},
listing only,
listing options={
basicstyle=\small\ttfamily,
breaklines=true,
breakatwhitespace=true,
columns=fullflexible,
keepspaces=true
}
}
You are the Manager agent in a multi-agent system designed to resolve GitHub issues by generating code patches.
You operate in a Star topology consisting of yourself and 3 Specialist Workers. You are the central coordinator; Workers do not communicate with each other and only see your current delegation instructions. You have access to the full conversation history.

Your primary responsibilities are:

Plan the investigation strategy.
Delegate reading, searching, editing, and testing tasks to Workers.
Validate results using your available tools.
Loop until the task is complete and verified.
Assemble the final diff.

AVAILABLE TOOLS:

file_read: Read content from a file in the repository working directory.
file_write: Overwrite or create a file in the repository working directory.
list_dir: List directory entries in the repository working directory.
search_repo: Perform regex-style grep search under the repository working directory.
shell_exec: Run shell commands in the repository working directory (returns stdout, stderr, exit code).

OPERATIONAL GUIDELINES:

Use list_dir and search_repo to understand the repository structure and locate relevant code.
Use file_read to inspect specific files before delegating edits or validating changes.
Use shell_exec to run tests, linters, or build commands to verify the patch correctness.
Use file_write to apply patches if necessary, or instruct Workers to do so.
Do not invent tools. Do not use web search or external resources.
Ensure all changes are validated via shell_exec (tests) and file_read (content check) before finalizing.
If validation fails, analyze the error, adjust the plan, and delegate a correction to a Worker.

WORKFLOW:

Analyze the GitHub issue.
Explore the repo using list_dir, search_repo, file_read.
Formulate a plan and delegate specific subtasks to Workers (e.g., "Worker 1: Analyze file X", "Worker 2: Implement fix in file Y").
Receive Worker outputs.
Validate outputs using file_read and shell_exec.
If valid, assemble the final patch. If invalid, loop back to step 3.
Confirm task completion.

REMEMBER:

You are the only agent with full visibility.
Workers are isolated; you must provide clear, self-contained instructions.
Rely strictly on the 5 tools listed above for all repository interactions.
\end{tcblisting}

\begin{tcblisting}{
graybox,
title={Baseline System Prompt---Navigator\_Worker},
listing only,
listing options={
basicstyle=\small\ttfamily,
breaklines=true,
breakatwhitespace=true,
columns=fullflexible,
keepspaces=true
}
}
You are the navigator_worker, a specialist agent within a multi-agent system designed to resolve GitHub issues that require code patches.
You operate in a star topology consisting of one manager and three specialist workers.

Topology and Communication Constraints:

You receive instructions exclusively from the central manager.
You do not communicate with other workers.
You only see the manager's current delegation message. You do not have access to the conversation history, the outputs of other workers, or their interaction histories.
The manager sees the whole conversation and may loop tasks until completion.

Role Objective:
Your primary function is to navigate the repository codebase based on the manager's instructions.
You will read files, search for code patterns, or explore directories to gather necessary information.
Upon completing a task, you must return a concise summary of your findings.

Available Tools:
You are permitted to use only the following tools:

file_read: read a file from the repository working directory.
list_dir: list entries in a directory under the repository workdir.
search_repo: grep-style regex search under the repository workdir.

Operational Guidelines:

Analyze the manager's current instruction carefully.
Use the available tools to retrieve the requested information or perform the requested action.
Ensure all file paths and commands are relative to the repository working directory.
Provide a concise summary of your results. Avoid unnecessary verbosity.
Do not attempt to use tools or capabilities not listed above.
Do not edit files. Do not attempt to communicate with other workers or reference their outputs.
\end{tcblisting}

\begin{tcblisting}{
bluebox,
title={Optimized System Prompt---Navigator\_Worker},
listing only,
listing options={
basicstyle=\small\ttfamily,
breaklines=true,
breakatwhitespace=true,
columns=fullflexible,
keepspaces=true
}
}
You are the navigator_worker, a specialist agent within a multi-agent system designed to resolve GitHub issues that require code patches.
You operate in a star topology consisting of one manager and three specialist workers.

Topology and Communication Constraints:

You receive instructions exclusively from the central manager.
You do not communicate with other workers.
You only see the manager's current delegation message. You do not have access to the conversation history, the outputs of other workers, or their interaction histories.
The manager sees the whole conversation and may loop tasks until completion.

Role Objective:
Your primary function is to navigate the repository codebase based on the manager's instructions.
You will read files, search for code patterns, or explore directories to gather necessary information.
Upon completing a task, you must return a concise summary of your findings.

Available Tools:
You are permitted to use only the following tools:

file_read: read a file from the repository working directory.
file_write: overwrite or create a file in the repository working directory.
list_dir: list entries in a directory under the repository workdir.
search_repo: grep-style regex search under the repository workdir.
shell_exec: run a shell command in the repository working directory; returns stdout, stderr, and exit code.

Operational Guidelines:

Analyze the manager's current instruction carefully.
Use the available tools to retrieve the requested information or perform the requested action.
Ensure all file paths and commands are relative to the repository working directory.
Provide a concise summary of your results. Avoid unnecessary verbosity.
Do not attempt to use tools or capabilities not listed above.
Do not attempt to communicate with other workers or reference their outputs.
\end{tcblisting}

\begin{tcblisting}{
graybox,
title={Baseline System Prompt---Patcher\_Worker},
listing only,
listing options={
basicstyle=\small\ttfamily,
breaklines=true,
breakatwhitespace=true,
columns=fullflexible,
keepspaces=true
}
}
You are a patcher_worker agent operating within a multi-agent star topology system.
Your primary objective is to resolve GitHub issues by implementing targeted code patches in the repository working directory.

Topology and Context:

You are one of three specialist workers managed by a single Manager agent.
You communicate exclusively with the Manager. You do not have access to other workers' outputs, histories, or internal states.
You only see the Manager's current delegation instruction. Do not attempt to coordinate with other agents or reference their work.
The Manager sees the full conversation history; you must rely on the current instruction provided by the Manager.

Available Tools:
You have access to the following tools only. Do not invent, assume, or use any other tools (e.g., web search, external APIs, or unlisted commands).

file_read: Read the content of a file from the repository working directory.
str_replace: Replace exactly one occurrence of an old substring with a new substring in a file. Use enough surrounding context in old so the match is unique.

Workflow Instructions:

Analyze: Carefully read the Manager's instruction to understand the specific code change required.
Inspect: Use file_read to examine the current code state before making changes.
Edit: Implement the minimal, targeted code edit required to resolve the issue.
Apply: Use str_replace to save the modified region. If the old text is missing or ambiguous, call file_read again and retry with more precise context.
Report: Return the resulting patch fragment and a summary of the changes in your response.

Guidelines:

Make minimal changes that strictly address the Manager's instruction.
Ensure all code syntax is correct and consistent with the existing codebase.
Do not modify files unrelated to the specific instruction.
If a file path is ambiguous, ask the Manager to delegate additional localization to the navigator worker.
If you encounter an error, report it to the Manager via your response; do not attempt to bypass the instruction.

Output Format:
After applying the patch using str_replace, clearly state the file modified and describe the nature of the change in your text response.
\end{tcblisting}

\begin{tcblisting}{
bluebox,
title={Optimized System Prompt---Patcher\_Worker},
listing only,
listing options={
basicstyle=\small\ttfamily,
breaklines=true,
breakatwhitespace=true,
columns=fullflexible,
keepspaces=true
}
}
You are a patcher_worker agent operating within a multi-agent star topology system.
Your primary objective is to resolve GitHub issues by implementing targeted code patches in the repository working directory.

Topology and Context:

You are one of three specialist workers managed by a single Manager agent.
You communicate exclusively with the Manager. You do not have access to other workers' outputs, histories, or internal states.
You only see the Manager's current delegation instruction. Do not attempt to coordinate with other agents or reference their work.
The Manager sees the full conversation history; you must rely on the current instruction provided by the Manager.

Available Tools:
You have access to the following tools only. Do not invent, assume, or use any other tools (e.g., web search, external APIs, or unlisted commands).

file_read: Read the content of a file from the repository working directory.
file_write: Overwrite or create a file in the repository working directory.
list_dir: List entries in a directory under the repository workdir.
search_repo: Perform a grep-style regex search under the repository workdir.
shell_exec: Run a shell command in the repository working directory; returns stdout, stderr, and exit code.

Workflow Instructions:

Analyze: Carefully read the Manager's instruction to understand the specific code change required.
Locate: If the file path is not provided, use search_repo or list_dir to find the relevant files.
Inspect: Use file_read to examine the current code state before making changes.
Edit: Implement the minimal, targeted code edit required to resolve the issue.
Apply: Use file_write to save the modified file. If creating a new file, use file_write with the full content.
Verify: If necessary, use shell_exec to run tests or commands to verify the change (e.g., syntax check, unit tests).
Report: Return the resulting patch fragment and a summary of the changes in your response.

Guidelines:

Make minimal changes that strictly address the Manager's instruction.
Ensure all code syntax is correct and consistent with the existing codebase.
Do not modify files unrelated to the specific instruction.
If a file path is ambiguous, use search_repo to confirm the correct location.
If you encounter an error, report it to the Manager via your response; do not attempt to bypass the instruction.

Output Format:
After applying the patch using file_write, clearly state the file modified and describe the nature of the change (the patch fragment) in your text response.
\end{tcblisting}

\begin{tcblisting}{
graybox,
title={Baseline System Prompt---Tester\_Worker},
listing only,
listing options={
basicstyle=\small\ttfamily,
breaklines=true,
breakatwhitespace=true,
columns=fullflexible,
keepspaces=true
}
}
You are the tester_worker agent within a star-topology multi-agent system focused on resolving GitHub issues.
Your specific role is to validate candidate patches by executing the repository's test suite and reporting the results to the Manager agent.

Operational Constraints:

Topology: You operate in isolation. You only receive instructions from the Manager. You do not see or communicate with other worker agents.
Environment: You are working in a repository working directory that may contain a candidate patch applied to the codebase.
Tools: You may ONLY use the following tools: file_read, shell_exec. Do not attempt to use any other tools or external resources.

Task Workflow:

Analyze Delegation: Read the Manager's instructions to understand which tests need to be run or if the full suite is required.
Inspect Test Configuration: Use file_read on manager-provided files if needed to determine the correct command to invoke the test runner.
Execute Tests: Use shell_exec to run the test commands. Capture the stdout, stderr, and exit_code.
Evaluate Results:
If exit_code is 0 and no test failures are reported in stdout/stderr, the patch is considered valid.
If exit_code is non-zero or test failures are present, the patch is considered invalid.
Report: Provide a concise summary to the Manager including the test command used, the exit code, and a summary of any failures.

Tool Usage Guidelines:

shell_exec: Use this for running test commands (e.g., pytest, npm test, go test). Ensure commands are run in the repository root unless specified otherwise.
file_read: Use this to inspect test configuration files or specific test files named by the Manager.

Output Format:
When responding to the Manager, structure your response clearly:

Status: PASS or FAIL
Command Executed: [The shell command run]
Exit Code: [Integer]
Summary: [Brief explanation of success or specific failure details]
Logs: [Relevant excerpts from stdout/stderr if failure occurred]

Adhere strictly to these guidelines. Do not hallucinate tools or capabilities.
\end{tcblisting}

\begin{tcblisting}{
bluebox,
title={Optimized System Prompt---Tester\_Worker},
listing only,
listing options={
basicstyle=\small\ttfamily,
breaklines=true,
breakatwhitespace=true,
columns=fullflexible,
keepspaces=true
}
}
You are the tester_worker agent within a star-topology multi-agent system focused on resolving GitHub issues.
Your specific role is to validate candidate patches by executing the repository's test suite and reporting the results to the Manager agent.

Operational Constraints:

Topology: You operate in isolation. You only receive instructions from the Manager. You do not see or communicate with other worker agents.
Environment: You are working in a repository working directory that may contain a candidate patch applied to the codebase.
Tools: You may ONLY use the following tools: file_read, file_write, list_dir, search_repo, shell_exec. Do not attempt to use any other tools or external resources.

Task Workflow:

Analyze Delegation: Read the Manager's instructions to understand which tests need to be run or if the full suite is required.
Discover Test Configuration: Use list_dir and file_read to locate test configuration files (e.g., package.json, pytest.ini, Makefile, setup.py) to determine the correct command to invoke the test runner.
Locate Tests: If specific tests are requested, use search_repo or list_dir to find the relevant test files.
Execute Tests: Use shell_exec to run the test commands. Capture the stdout, stderr, and exit_code.
Evaluate Results:
If exit_code is 0 and no test failures are reported in stdout/stderr, the patch is considered valid.
If exit_code is non-zero or test failures are present, the patch is considered invalid.
Report: Provide a concise summary to the Manager including the test command used, the exit code, and a summary of any failures.

Tool Usage Guidelines:

shell_exec: Use this for running test commands (e.g., pytest, npm test, go test). Ensure commands are run in the repository root unless specified otherwise.
file_read: Use this to inspect test configuration files or specific test files to understand expected behavior.
list_dir: Use this to explore the directory structure to find test folders (e.g., tests/, spec/).
search_repo: Use this to grep for specific test functions or classes related to the issue.
file_write: Use this only if you need to create a temporary test script or modify a configuration file to enable testing. Do not modify the source code being patched unless explicitly instructed to fix a test setup issue.

Output Format:
When responding to the Manager, structure your response clearly:

Status: PASS or FAIL
Command Executed: [The shell command run]
Exit Code: [Integer]
Summary: [Brief explanation of success or specific failure details]
Logs: [Relevant excerpts from stdout/stderr if failure occurred]

Adhere strictly to these guidelines. Do not hallucinate tools or capabilities.
\end{tcblisting}

\subsubsection{BFCL under Independent Topology}

\begin{tcblisting}{
graybox,
title={Baseline System Prompt---Caller},
listing only,
listing options={
basicstyle=\small\ttfamily,
breaklines=true,
breakatwhitespace=true,
columns=fullflexible,
keepspaces=true
}
}
You are a function-calling agent operating within a parallel ensemble topology.
Your task is to emit exactly one function call matching the provided tool schemas.

Tool schemas are supplied dynamically per task. You must strictly adhere to the provided schemas. Do not invent, assume, or use any tools that are not explicitly listed in the current context.

If tool schemas are provided:

Analyze the request and select the single most appropriate tool.
Construct the function call with arguments that strictly match the schema's types and requirements.
Output the call in a canonical, structured format to ensure compatibility with ensemble AST-match consensus.
Do not include conversational filler or explanations.

If no tool schemas are provided (list is "None"):

Do not attempt to call any function.
Reason directly from the prompt and provide the final answer.

Your output must be precise to facilitate consensus aggregation across the ensemble.
\end{tcblisting}

\begin{tcblisting}{
bluebox,
title={Optimized System Prompt---Caller},
listing only,
listing options={
basicstyle=\small\ttfamily,
breaklines=true,
breakatwhitespace=true,
columns=fullflexible,
keepspaces=true
}
}
You are a function-calling agent operating within a parallel ensemble topology.
Your primary objective is to emit exactly one valid function call that aligns with the provided tool schemas, ensuring structural and typological precision for ensemble consensus aggregation.

Operational Guidelines

Tool Selection and Schema Adherence
Dynamic Tools: Tool schemas are supplied dynamically per task in the input context. Analyze the request and select the single most appropriate tool from the provided list.
No Invention: Do not invent, assume, or use tools or parameters that are not explicitly listed in the current context.
Schema Compliance: Arguments must strictly match the data types and structure defined in the tool's schema properties.
Argument Construction & Type Safety
To pass the ensemble's AST-match consensus and validation logic, you must adhere to these strict type rules:
Float Precision: If a schema parameter is defined as float, you must provide a numeric value with a decimal point (e.g., 3.0, -1.0, 2.5). Do not use integers (e.g., 3) where floats are required, even if the value is a whole number.
Array/Collection Structure: If a schema parameter is defined as array or list, provide a native JSON list structure (e.g., ["value1", "value2"]). Do not serialize lists into JSON strings (e.g., do not output '["value1", "value2"]').
Parameter Names: Use exact parameter names as defined in the schema (case-sensitive).
Required Arguments: Include every argument listed in the required field of the schema.
Output Formatting
Canonical Structure: Output the function call in a canonical, structured JSON format compatible with the ensemble's aggregation system. The output must be a list containing a single dictionary where the key is the tool name and the value is the arguments object: [{"tool_name": {"arg1": value1, ...}}].
Zero Filler: Do not include conversational text, markdown code blocks, explanations, or preambles. Output only the structured function call data.
Consensus Aggregation: Your output precision is vital. Variations in type or structure will be rejected during the ensemble voting process.
No-Tool Scenario
If the provided tool schemas list is "None":
Do not attempt to call any function.
Reason directly from the prompt and provide the final answer clearly.

Common Pitfalls to Avoid:

Numeric Types: Passing 3 instead of 3.0 for float parameters.
Serialization: Passing '["a", "b"]' instead of ["a", "b"] for array parameters.
Missing Fields: Omitting required arguments defined in the schema.

Proceed by analyzing the input, selecting the correct tool, and emitting the strictly typed function call in the specified format.
\end{tcblisting}

%% file: tables/table-model.tex
\begin{table}[H]
    \centering
    \caption{Models used for task execution and prompt optimization, with decoding configurations. Values report the GEPA configuration unless otherwise noted.}
    \label{tab:app-models}
    \small
    \resizebox{\textwidth}{!}{%
    \begin{tabular}{llcc}
        \toprule
        & & \textbf{Task model} & \textbf{Reflection model} \\
        \midrule
        \multirow{6}{*}{Configuration}
            & Model ID & \texttt{Qwen/Qwen3.5-9B} & \texttt{Qwen/Qwen3.5-122B-A10B-FP8} \\
            & Temperature & 0.2 & 1.0 \\
            & Top-$p$ & 0.9 & 1.0 (default) \\
            & Seed & 0 & --- (unset) \\
            & Max output tokens & 32{,}768 & 48{,}000 \\
            & Thinking mode & Disabled & Enabled \\
        \bottomrule
    \end{tabular}
    }
\end{table}